\documentclass[journal]{IEEEtran}
\usepackage{graphicx}
\usepackage{cite}
\usepackage[bookmarks=false]{hyperref}
\usepackage{multirow}
\usepackage{enumerate}
\usepackage{bm}
\usepackage{subfigure}
\usepackage{booktabs}
\usepackage{blindtext}
\usepackage[T1]{fontenc}
\usepackage{mathptmx} 
\DeclareMathAlphabet{\mathcal}{OMS}{cmsy}{m}{n}

\usepackage{times} 
\usepackage{amsmath} 
\usepackage{amssymb}  
\usepackage[ruled,vlined]{algorithm2e}
\usepackage{array}
\newcommand{\PreserveBackslash}[1]{\let\temp=\\#1\let\\=\temp}
\newcolumntype{C}[1]{>{\PreserveBackslash\centering}p{#1}}
\newcolumntype{R}[1]{>{\PreserveBackslash\raggedleft}p{#1}}
\newcolumntype{L}[1]{>{\PreserveBackslash\raggedright}p{#1}}

\usepackage{pifont}
\usepackage[dvipsnames]{xcolor}

\IEEEaftertitletext{\vspace{-2.0\baselineskip}}
\begin{document}
\title{VTGNet: A Vision-based Trajectory Generation Network for Autonomous Vehicles \\ in Urban Environments
}

\author{Peide~Cai,
        Yuxiang~Sun,
        Hengli~Wang,
        and~Ming~Liu,~\IEEEmembership{Senior Member,~IEEE}

\thanks{Peide Cai, Yuxiang Sun, Hengli Wang and Ming Liu are with the Department of Electronic and Computer Engineering, Hong Kong University of Science and Technology, Clear Water Bay, Kowloon, Hong Kong SAR, China (e-mail: pcaiaa@connect.ust.hk; sun.yuxiang@outlook.com; hwangdf@connect.ust.hk; eelium@ust.hk) \textit{(Corresponding author: Ming Liu.)} }
}

\maketitle

\begin{abstract}
Traditional methods for autonomous driving are implemented with many building blocks from perception, planning and control, making them difficult to generalize to varied scenarios due to complex assumptions and interdependencies. Recently, the end-to-end driving method has emerged, which performs well and generalizes to new environments by directly learning from export-provided data. However, many existing methods on this topic neglect to check the confidence of the driving actions and the ability to recover from driving mistakes. In this paper, we develop an uncertainty-aware end-to-end trajectory generation method based on imitation learning. It can extract spatiotemporal features from the front-view camera images for scene understanding, and then generate collision-free trajectories several seconds into the future. The experimental results suggest that under various weather and lighting conditions, our network can reliably generate trajectories in different urban environments, such as turning at intersections and slowing down for collision avoidance. Furthermore, closed-loop driving tests suggest that the proposed method achieves better cross-scene/platform driving results than the state-of-the-art (SOTA) end-to-end control method, where our model can recover from off-center and off-orientation errors and capture 80\% of dangerous cases with high uncertainty estimations.
\end{abstract}


\section{Introduction}

\IEEEPARstart{F}{rom} a global perspective, approximately 1.3 million people die yearly due to road traffic\cite{global_report1}, and nearly 94\% of these are related to human driving errors\cite{human_error}. Autonomous vehicles (AVs) may play an essential role in reducing this number, whilst alleviating traffic congestion, cutting down air pollution and reducing energy consumption in transportation by as much as 90\% \cite{global_report3}.

\begin{figure}[t]
        \centering
        \includegraphics[width = \columnwidth]{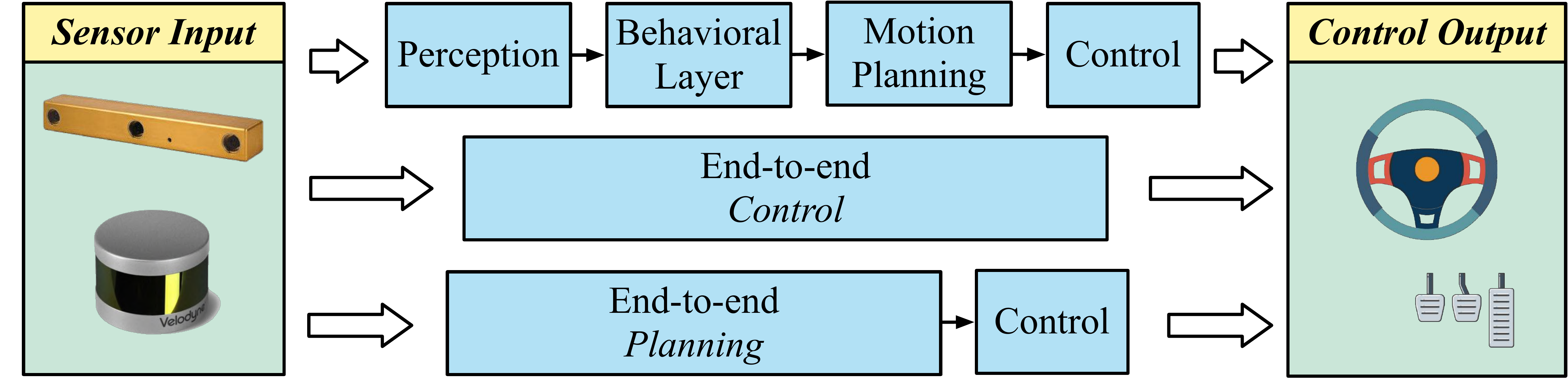}
        \caption{Different approaches for trajectory planning and decision-making for autonomous vehicles.}
        \label{fig:pipelines}
\end{figure}

In order to achieve autonomous driving, vehicles need to perceive and understand their surroundings\cite{fan2020sne}, then use the extracted information to generate collision-free trajectories to the goal position. However, due to the highly complex environment and an inability to test the system in a wide variety of scenarios\cite{kuutti2020survey}, achieving universal autonomous driving is still a challenge, especially in real-world urban environments where trajectory generation is a crucial task. Within this framework, the solution trajectory can be represented as a time-parametrized function, $\pi(t):[0, T] \rightarrow \mathcal{X}$, where $T$ is the planning horizon and $\mathcal{X}$ is the configuration space of the vehicle\cite{survey_trajectory}. Methods for planning and decision-making for AVs can be divided into three main categories\cite{schwarting2018planning, cai2019vision}: traditional sequential planning, end-to-end control and end-to-end planning, which are illustrated in Fig. \ref{fig:pipelines}.

The traditional approaches are usually structured as a pipeline of separate components (e.g., perception, planning and control) linking sensory inputs to actuator outputs\cite{mcallister2017concrete}. However, there exist some major disadvantages of this approach. 1) Each component needs to be individually specified and tuned, which is difficult to generalize to varied scenarios due to their complex interdependencies\cite{learn_to_drive}. 2) The cost functions in most works have to be carefully designed with complicated parameters. 3) It is prone to error propagation. The accumulated error among unstable modules may lead to false or missed alarms. For example, in the unfortunate Tesla accident in 2016, an error in the perception module, misclassification of a white trailer as sky, propagated down the pipeline until failure\cite{yurtsever2019survey}.

Recently, a new paradigm based on deep learning, which integrates perception, planning and control has achieved impressive results on vehicle navigation\cite{eed1,eed2,cai2020probabilistic, condition-eed, bddv, Mller2018DrivingPT, codevilla2019exploring}. Called \textit{end-to-end control}, this method formulates the driving problem as learning a mapping from perceptual inputs (e.g., RGB images) directly to vehicle control outputs (e.g., steering angle and throttle). Another similar paradigm is \textit{end-to-end planning} \cite{cai2019vision, thesis, r2p2, waymo}, which maps observations into future trajectories. Deep learning enables these systems to perform well and generalize to new environments by self-optimising (learning) its behaviour from data, allowing freedom from tuning rule-based parameters in all foreseeable scenarios. However, these works have several shortcomings. 1) The driving ability of recovering from off-center and off-orientation errors has not been considered or validated, which is critical for real-world deployment. 2) Most works only evaluate their models in simple environments with limited complexity in terms of dynamic obstacles and weather/lighting conditions. 3) Additionally, the generalization performance on different types of vehicles with changing physical properties has seldom been evaluated. 4) The neglected uncertainty of the generated driving actions or trajectories prevents the users from checking the correctness of the network output, which is crucial in safety-critical driving tasks. 5) Finally, many works heavily rely on post-processed detailed environmental maps for decision making \cite{waymo, r2p2}, which are costly to create and transfer to new scenarios.

Towards addressing the above problems, we follow the more generic end-to-end planning paradigm and propose VTGNet, an uncertainty-aware vision-based trajectory generation network for AVs. This paper is an extended version of \cite{cai2019vision} and the main contributions are summarized as follows.

\begin{enumerate}
\item We propose an end-to-end driving network for AVs based on imitation learning, which directly takes as input the raw camera images rather than highly-engineered environmental maps for trajectory generation.

\item We propose a new benchmark \textit{AddNoise} to examine the driving ability to recover from off-center and off-orientation errors.

\item We compare our VTGNet with different baselines and show its superiority under various weather/lighting conditions (e.g., snowy, rainy and foggy) both on a large-scale driving dataset and in a high-fidelity driving simulator with dynamic obstacles.

\item We demonstrate the cross-scene/platform and error recovery driving ability of VTGNet on the \textit{AddNoise} benchmark.

\item We validate the effectiveness of the estimated uncertainty by ablation studies and closed-loop driving tests.

\item We open source our code and collected synthetic driving dataset for future studies on autonomous driving\footnote{\url{https://github.com/caipeide/VTGNet}}.

\end{enumerate}

\section{Related Work}

\textbf{End-to-end control.}
ALVINN (autonomous land vehicle in a neural network), proposed by Pomerleau\cite{alvinn} in 1989, is a pioneer attempt to use the neural network for autonomous driving. Due to the composition of a limited number of neural network layers, this method only works in very simple scenarios. In 2016, with the development of convolutional neural networks (CNNs) and computing powers of GPUs, Bojarski \textit{et al.}\cite{eed1} developed a more advanced driving model named DAVE-2. It achieves autonomous lane following in relatively simple real-world scenarios, such as flat or barrier-free roads, in which the front-view camera is used to stream the images and transmit them into CNNs to compute steering commands. Follow-up works include \cite{bddv}, \cite{eed2} and \cite{wu2019end}. However, these works only target \textit{lane-following} tasks, and since they only consider the camera input for decision making, a wrong turn may be taken at intersections for the lack of high-level navigation commands. Moreover, only simple environments with low-level complexity are considered in these works.

\textbf{Incorporating intentions into driving networks.} In order to resolve the ambiguities at intersections, driving models proposed by Codevilla \textit{et al.}\cite{condition-eed, codevilla2019exploring} have been designed to receive not only the perceptual inputs but also high-level driving intentions (i.e., \textit{keep straight} and \textit{turn left}). In this way, the network becomes more controllable. Similarly, Cai \textit{et al.} \cite{cai2020high} realize high-speed autonomous drifting in racing scenarios guided by route information with deep reinforcement learning. However, the driving policy is only demonstrated in static enviroments. Hecker \textit{et al.} \cite{eth-route} trained a driving model to use 360-degree camera images equipped with GPS-based route information to predict future steering and speed controls. However, this work was only evaluated \textit{offline} on a pre-collected dataset. Closed-loop \textit{online} driving abilities such as slowing down for pedestrians/vehicles are not presented.

\textbf{End-to-end planning using temporal information.}
To make the driving models more generic, new methods have been recently proposed for end-to-end trajectory planning \cite{cai2019vision, thesis, r2p2, waymo}. These methods commonly use recurrent neural networks (RNNs) to handle temporal sequential information (e.g., images). For example, Bergqvist \textit{et al.} \cite{thesis} designed and tested several path planning networks with various types of input including gray-scale images and ego-motions. The results showed that the path generated by long short-term memory (LSTM) or CNN-LSTM is smooth and feasible in many situations. However, this work only considered lane following tasks in simple areas. To handle more complex driving situations, Rhinehart \textit{et al.} \cite{r2p2} consider the future trajectories as a distribution conditioned on post-processed top-view feature maps rather than raw camera images, from which a set of possible paths can be sampled. Similarly, Bansal \textit{et al.} \cite{waymo} proposed to use top-view environmental representations to generate trajectories with RNNs. However, the detailed top-down views used in these two works are expensive to create, maintain and transfer, and they rely on prebuilt high-resolution maps of the driving areas.

Our work corresponds to the idea of end-to-end planning but differs from the other works in three main aspects: 1) Our method is free from detailed top-view feature/environmental maps and is directly based on highly accessible raw camera images for scene understanding and trajectory generation. 2) The history sequential input is redundant and may contribute differently to the model\cite{yang2018scene}, which, however, is neglected by prior RNN-based driving models. In the spirit of advancements in text translation\cite{bahdanau2014neural} and graph representation learning\cite{zhang2019heterogeneous}, we design a \textit{self-attention} LSTM module to better use the temporal information. 3) Our method is extensively evaluated both on a large-scale driving dataset (offline, open-loop) and in a high-fidelity driving simulator (online, closed-loop).

\begin{figure*}[t]
        \centering
        \includegraphics[width = 2\columnwidth]{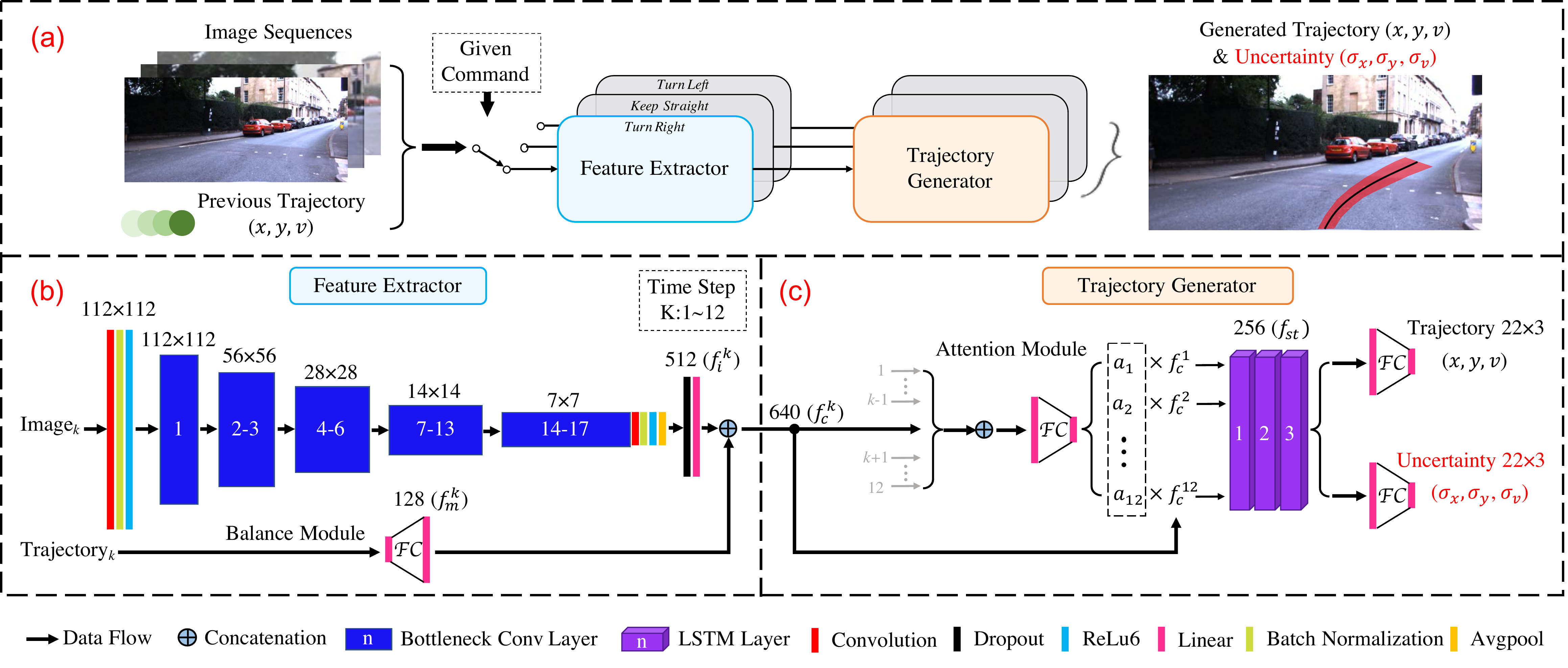}
        \caption{Architecture of the proposed VTGNet, which consists of a feature extractor and a trajectory generator. VTGNet is composed of three sub-networks that are selectively activated for three common tasks: \textit{turn left}, \textit{turn right} and \textit{keep straight}. MobileNet-V2 is used as the feature extractor with 17 bottleneck convolutional layers. LSTM is used to process the spatiotemporal information. We also adopt a self-attention mechanism to measure the importance of the past sequential information before feeding it into the LSTM. The output of VTGNet is two-branched: 1) A vector of size $22\times 3$ indicating the trajectory in the future 22 frames (velocity and $x,y$ positions in the body frame), and 2) a vector of size $22\times 3$ indicating the uncertainty of the generated trajectory.}
        \label{fig:vtgnet}
\end{figure*}

\textbf{Learning uncertainty-aware navigation policies.} Recently, several methods have been proposed for uncertainty estimation in deep neural networks. Kendall \textit{et al.}\cite{Kendall2017WhatUD} proposed that there are two types of uncertainties: the \textit{aleatoric} uncertainty and the \textit{epistemic} uncertainty. The \textit{epistemic} uncertainty is the model uncertainty, which can be reduced by adding sufficient data. \textit{Aleatoric} uncertainty, also known as data uncertainty, captures the uncertainty from the observed input data. Tai \textit{et al.}\cite{Tai2019VisualbasedAD} accomplished uncertainty-aware visual-based navigation by using data uncertainty to choose the safest action among multiple network outputs. Fan \textit{et al.}\cite{Fan2019LearningRB} also used the data uncertainty to achieve resilient behaviors in prior unknown environments for indoor navigation. Similar to these works, we show performance improvements by explicitly modeling the data uncertainty into deep networks. Differently, we find this method is still insufficient for hard-coded safety on autonomous driving, which opens possible avenues for future research in this area.

\section{Methodology}

\subsection{Network Architecture}

The overall architecture of the proposed VTGNet is shown in Fig. \ref{fig:vtgnet}-(a). At each time step $t$, the inputs to the network are camera images $\mathcal{I}_t = \{\mathcal{I}^1,...,\mathcal{I}^{12}\}$ and the movement information of the vehicle $\bm{m}_t = \{\bm{m}^1,...,\bm{m}^{12}\}$ in the past 12 frames. The output of the network is the predicted trajectory $\mathcal{T}_t$ and corresponding uncertainty $\sigma_t$, which are conditioned on the high-level driving command $\bm{c}_t$ and observations. The preview horizon of the generated trajectory is set to 3 s (i.e., 22 frames), which is double the average human perception-brake reaction time (RT) of 1.5 s to stop a vehicle \cite{time_ref}. Therefore, the generated trajectory can be denoted as $\mathcal{T}_t = \{\mathcal{T}^1,...,\mathcal{T}^{22}\}$. Here, $\bm{m}^k \in \mathbb{R}^3$ and $\mathcal{T}^k \in \mathbb{R}^3$ both contain the velocity and $x,y$ positions in the current body frame. Similarly, the trajectory uncertainty is denoted as $\sigma_t = \{\sigma^1,...,\sigma^{22}\}$, where $\sigma^k \in \mathbb{R}^3$ contains uncertainties of the velocity and $x,y$ positions.

For implementation, we construct three sub-networks that can be selected by the given command $\bm{c}_t$ to conduct different tasks, i.e., \textit{turn left}, \textit{turn right} and \textit{keep straight}, similar to other works \cite{condition-eed, codevilla2019exploring}. In practice, as in \cite{Kendall2017WhatUD}, we train the network to predict the log variance log$\sigma^2$ because it is more numerically stable than regressing the variance.

\subsubsection{The Feature Extractor} Once a specific branch is chosen by a high-level command, the images $\mathcal{I}_t$ are first processed separately and in turn by an image module $F_i$ implemented with CNNs, which extracts a feature vector $f_i \in \mathbb{R}^{512}$ from each image. We use MobileNet-V2\cite{mobilenetv2} as $F_i$ to extract visual features. It consists of 17 bottleneck convolutional blocks, and each block uses the depthwise separable convolutions rather than a fully convolutional operator. We refer readers to \cite{mobilenetv2} for more details of this module. Then, a balance module $F_b$ implemented with fully connected (FC) layers expands the dimension of each history movement vector $\bm{m}^k$ from 3 to 128 ($f_m^k$) to balance the influence of the vision and motion feature vectors, for which we draw on the experience of \cite{intention}. The outputs of these two modules are then concatenated together at every history time step into combined vectors of length 640, represented by
\begin{equation}
	f_c^k = < F_i(\mathcal{I}^k),\  F_b(\bm{m}^k) >,\ 1\leq k\leq 12,
\end{equation}
where $<\cdot>$ represents the concatenation operation,  $F_i(\mathcal{I}^k)$ is the output of the image module for the k-th image, $F_b(\bm{m}^k)$ is the output of the balance module for the k-th motion state vector, and $f_c^k$ is the k-th combined vector. 

\subsubsection{The Trajectory Generator} To better use the sequential information, the set of $f_c = \{ f_c^1,...,f_c^{12} \}$ is first concatenated and processed by FC layers to generate attentions $a_1,...,a_{12}$ to measure the relative importance of the past information. We use \textit{softmax} to activate the last layer so that $\sum_1^{12}a_k=1$. Then, we feed the modulated features $\{a_1f_c^1,...,a_{12}f_c^{12}\}$ into a three-layer LSTM to generate the spatiotemporal feature $f_{st} \in \mathbb{R}^{256}$ of the surroundings. It is further compressed by 2 FC layers to get two vectors, both of size $22 \times 3$, which represent the trajectory $\mathcal{T}_t$ and uncertainty $\sigma_t$ generated at time step $t$ for the future 22 frames.

\subsubsection{Loss Function}
As the name suggests, the basic idea of imitation learning is to train a network that mimics human behaviors, which can be solved with supervised learning. Let $\mathcal{T}_t^{\prime}$ denote the expert-provided trajectory in the training dataset at sample time \textit{t}, and $\theta$ denote the learnable parameters of the network $F$. Then, the optimal parameters $\theta^{\ast}$ can be trained by minimizing the average prediction loss:
\begin{equation}
    \theta^{\ast} = \mathop{\arg\min}_{\theta} \sum_t \mathcal{L}\left(F\left(\mathcal{I}_t, \bm{m}_t, \bm{c}_t; \theta \right), \mathcal{T}_t^{\prime} \right),
\end{equation}
where $\mathcal{L}$ is the per-sample loss at sample \textit{t}. We follow the training method for data uncertainty in \cite{Kendall2017WhatUD} and define $\mathcal{L}$ as
\begin{equation}
\begin{split}
    \mathcal{L} = \frac{\left \Vert \mathcal{T}_t - \mathcal{T}_t^{\prime} \right \Vert_2^2}{2\sigma_t^2} + \frac{1}{2}\text{log}\sigma_t^2.
\end{split}
\end{equation}

\begin{figure}[t]
        \centering
        \includegraphics[width = \columnwidth]{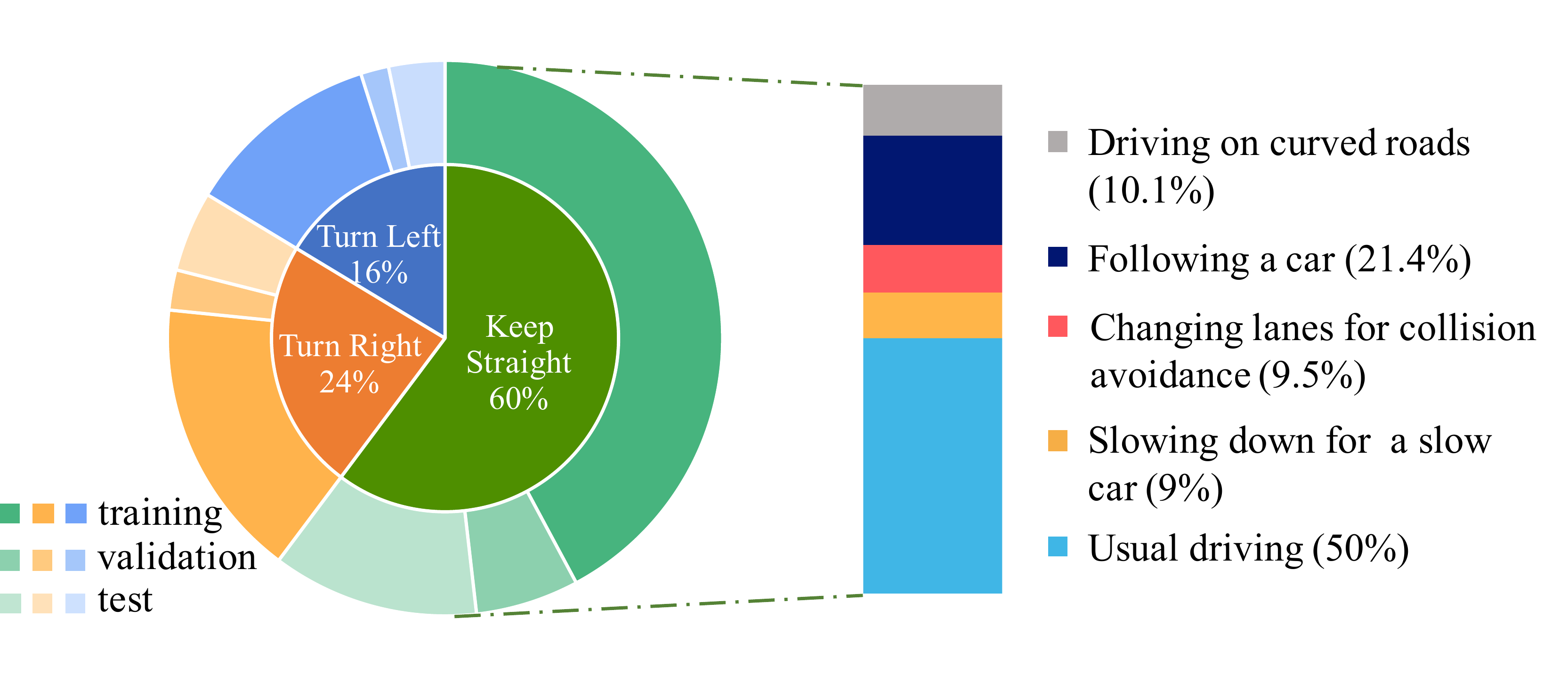}
        \setlength{\abovecaptionskip}{-0.5cm}
        \caption{Dataset distribution and splitting. For each situation, the split ratio of training, validation and test is 7:1:2. The staked column bar on the right shows the distribution of data in the \textit{keeping straight} situation, where we have balanced the portion for usual and unusual driving cases. The turning situation does not contain significant interaction with the road agents.}
        \label{plot:dataset_distribution}
\end{figure}

\begin{table}[t]
        \renewcommand{\arraystretch}{1.3}
        \caption{The Environmental Distributions of Our Dataset}
        \label{tab:dataset}
        \centering
        \begin{tabular}{l | c  c c}
        \toprule[1pt]
        Environments &  \# Turn Left & \# Turn Right  & \# Keep Straight \\
        \midrule
        Sun& 3705 & 5355 & 7921 \\
        Rain & 3369 & 4401 & 16548\\
        Snow &1280 & 3975 & 6089 \\
        Dusk &1059 & 1194 & 4836 \\
        Night &2339 & 2705 & 10930 \\   
        Overcast & 2733 & 3103 & 7016\\
        \bottomrule[1pt]
        \end{tabular}
\end{table}

\vspace{-0.5cm}
\subsection{Dataset Generation and Analysis}
\label{subsec:dataset}

We build our dataset from the RobotCar dataset, which was recorded in dynamic urban areas from May 6, 2014 to November 13, 2015. This dataset captures scenarios with various weather and lighting conditions, along with some long-term changes such as construction and roadworks. The front-view images are recorded by a Point Grey Bumblebee XB3 camera on top of the vehicle. In addition, the ground-truth positions and velocities are acquired from the fused GPS/inertial solution at a frequency of 50 Hz. 

\subsubsection{Data Generation}
From the RobotCar dataset, we aim to extract both camera images and movement information to train and test VTGNet. Therefore, the data without GPS information is first \textit{filtered} out and a total of 29 driving routes are collected. Another issue that needs to be considered is \textit{data balance}. Specificallly, we need to balance the portion for different driving cases (e.g., car following, slowing down for collision avoidance, etc.) and the distribution of driving samples in varied environmental conditions. The final dataset distribution is shown in Fig. \ref{plot:dataset_distribution} and in Table \ref{tab:dataset}.

This data is then \textit{reconstructed} for training and testing in this work, where we need to equip every image with the trajectory of the ego-vehicle in the past 1.5 s and the future 3 s. For implementation, we first interpolate the ground-truth universal transverse mercator (UTM) position and velocity series to the image timestamps, which are recorded at 15 Hz, and then convert their coordinates to the vehicle body frame. After this, we manually label every image with a corresponding intention command indicating the driving direction of the vehicle based on the ground-truth trajectories. For example, when the ego-vehicle approaches an intersection and plans to take a right turn, we change the command from \textit{keep straight} to \textit{turn right}. When the steering is completed, we recover the command to \textit{keep straight}. Finally, we crop the raw images to the shape of (1247, 384) by removing the sky and hood on the top/bottom of the images because these areas are less informative for scene understanding. We believe that the learning process is more efficient by doing so.

\begin{figure}[t]
        \centering
        \includegraphics[width = 0.7\columnwidth]{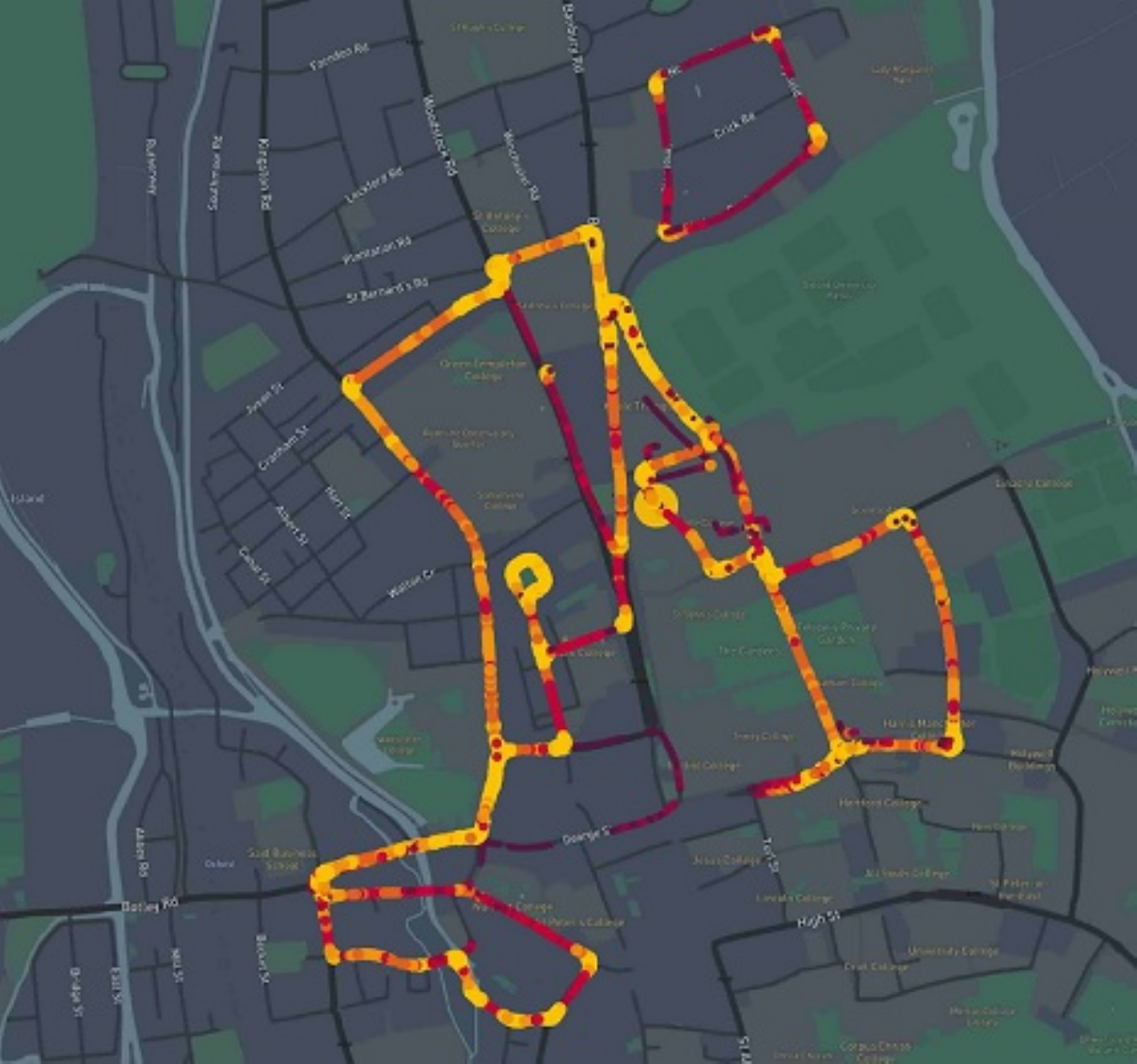}
        \caption{Dataset distribution map. We plot and cluster the locations of all the samples from our dataset. Clusters with more samples have a brighter color and are larger in size.}
        \label{fig:distribution_map}
\end{figure}

\subsubsection{Data Analysis} Our final dataset contains 88,558 images of driving sequences in Oxford for 61.52 km, which covers 6 environmental conditions:\textit{ overcast, sun, rain, snow, dusk} and \textit{night}. Each environment has a special style of visual appearance. For example, in the \textit{rain} scenario, the roads are often covered by fallen leaves, and sometimes raindrops on the camera lens cause blurry areas in the image. In the \textit{night} scenario, the shutter speed is much slower than it is in the daytime, which can lead to motion artifacts around objects when the vehicle moves. In the \textit{snow} scenario, the undrivable areas are often covered by white snow, contrasting sharply with the drivable areas. Sample visual appearances can be seen in Fig. \ref{fig:qualitative}. In addition, to visualize how the dataset distributes geologically, we plot the locations on an aerial map in Fig. \ref{fig:distribution_map}.

\begin{table*}[t]
        \renewcommand{\arraystretch}{1.3}
        \caption{\textsl{Models}: The Quantitative Evaluation Results for Different Models. \textsl{Environments}: The Evaluation Results of VTGNet in Different Weather/Lighting Conditions. Smaller Numbers Are Better. The Bold Font Highlights the Best Result in Each Column.}
        \label{tab:quantitative}
        \centering
        \begin{tabular}{l C{1.15cm} C{1.15cm} C{1.15cm} C{1.15cm} C{1.15cm} C{1.15cm} C{1.15cm}}
        \toprule[1pt]
        \multirow{2}{*}{}
        
        & \textsl{Accel} & $\mathcal{E}_{v}$ & $\mathcal{E}_{acc}$ & $\mathcal{E}_{ad}$ & $\mathcal{E}_{x}$ & $\mathcal{E}_{y}$ & $\mathcal{E}_{fd}$\\
        
        \textsl{Models} & $(m/s^2)$ & $(m/s)$ & $(m/s^2)$ & $(m)$ & $(m)$ & $(m)$ & $(m)$\\
        \midrule
        \texttt{CNN-FC}\cite{bddv} &\textbf{0.181} &1.104 &0.313 &1.621 &0.268 &1.545 &3.271\\
        \texttt{CNN-LSTM}\cite{bddv} &0.488 &0.995 &0.564 &1.461 &0.269 &1.380 &2.907 \\
        \texttt{CNNState-FC}\cite{thesis} &0.326 &0.477 &0.399 &0.628 &0.242 &0.521 &1.444\\
        \texttt{VTGNet} \textit{(ours)}&0.325 &\textbf{0.289} &\textbf{0.311} &\textbf{0.426} &\textbf{0.188} &\textbf{0.335} &\textbf{1.036}\\
        
        \midrule
        \textsl{Environments}\\
        \midrule
        \texttt{\textit{Sun}} &0.402 &0.286 &0.339 &0.470 &0.193 &0.372 &1.120    \\
        \texttt{\textit{Rain}} &0.294 &0.270 &0.299 &0.409 &0.172 &0.329 &0.962   \\
        \texttt{\textit{Dusk}} &0.350 &0.359 &0.353 &0.421 &0.185 &0.332 &1.095  \\
        \texttt{\textit{Night}} &0.318 &0.310 &0.308 &0.414 &0.205 &0.310 &1.040 \\
        \texttt{\textit{Overcast}} &0.337 &0.311 &0.336 &0.508 &0.204 &0.418 &1.226   \\
        \texttt{\textit{Snow}}  &\textbf{0.262} &\textbf{0.236} &\textbf{0.247} &\textbf{0.329} &\textbf{0.171} &\textbf{0.237} &\textbf{0.818} \\
        
        \bottomrule[1pt]
        \end{tabular}
\end{table*}

\section{Experiments on the RobotCar Dataset}
\label{sec:experiments_robotcar}

\subsection{Training Details and Baselines}
\begin{figure}[t]
        \centering
        \includegraphics[width = \columnwidth]{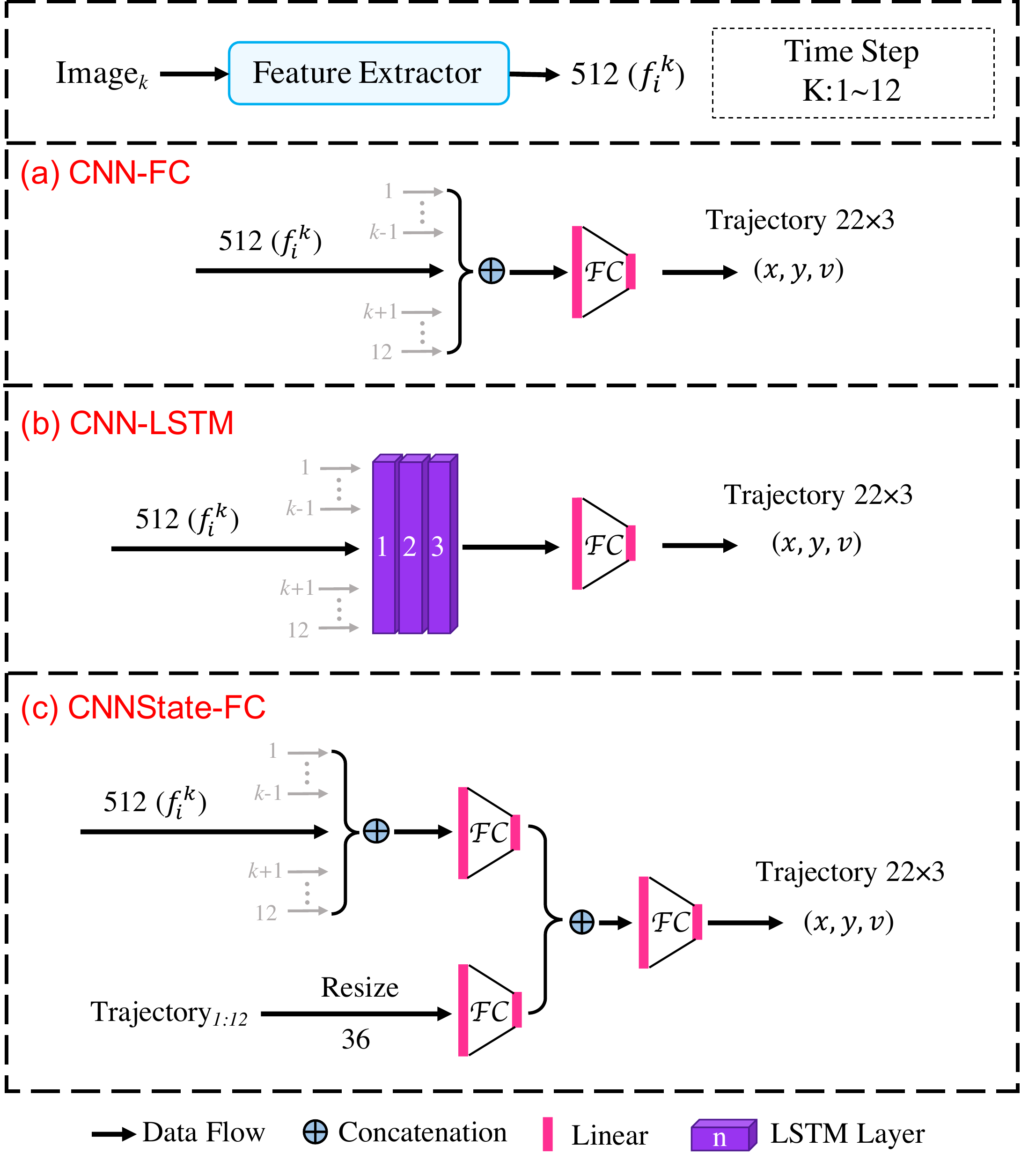}
        \caption{Different baselines for comparison. The feature extractor for these networks is the same as that in our proposed \texttt{VTGNet} shown in Fig. \ref{fig:vtgnet}}
        \label{fig:baseline_structures}
\end{figure}

We implement the proposed \texttt{VTGNet} with Pytorch, and train it on the dataset introduced in Section \ref{sec:experiments_robotcar} with the NVIDIA 1080Ti graphics card. The split ratio of training, validation and test is set to 7:1:2. We use the Adam optimizer\cite{adam} with an initial learning rate of 0.0001 and batch size of 15. The network is trained to converge when no further decrease in the validation loss can be observed. For comparison, we also train and fine-tune three other vision-based end-to-end trajectory planning baselines on the same training set. They are introduced as follows, and are shown in Fig. \ref{fig:baseline_structures}.

\texttt{CNN-FC}. This network takes as input the image sequences. The CNNs first extract the visual features in the past 12 frames (1.5 s), and then these features are concatenated together to be compressed by FC layers. This method follows the idea of \textit{TCNN} introduced in \cite{bddv}.
	    
\texttt{CNN-LSTM}. This network takes as input the image sequences. The extracted features in the past 12 frames are processed by a three-layer LSTM block into a vector $\in \mathbb{R}^{512}$. The vector is then compressed by FC layers. This method follows the idea of \textit{CNN-LSTM} introduced in \cite{bddv}.
        
\texttt{CNNState-FC}. This network takes as input both image and movement sequences. The extracted features are concatenated and compressed with FC layers. This method follows the idea of \textit{Merging Model} introduced in \cite{thesis}.

We adopt 7 metrics to evaluate the performance of different networks. These are the average values computed over the entire test dataset.
\begin{itemize}
      \item \textsl{Accel} measures the smoothness of the generated trajectory, which is the average acceleration of the preview time. The lower its values, the smoother and more comfortable the corresponding trajectories.
      \item $\mathcal{E}_{v}$ and $\mathcal{E}_{acc}$ measures the mean velocity and acceleration error, respectively, in the preview horizon of a trajectory.
      \item $\mathcal{E}_{ad}$ represents the average displacement error\cite{metrics}. It is the L2 distance between the generated and the ground-truth trajectory.
      
      \item $\mathcal{E}_{x}$ and $\mathcal{E}_{y}$ represents the lateral and longitudinal error, respectively, of a trajectory.
      
      \item $\mathcal{E}_{fd}$ is the final displacement error\cite{metrics}, which means the L2 distance between the final waypoints of the generated and ground-truth trajectories.
      
\end{itemize}

\begin{figure}[t] 
        \centering
        \includegraphics[width=\columnwidth]{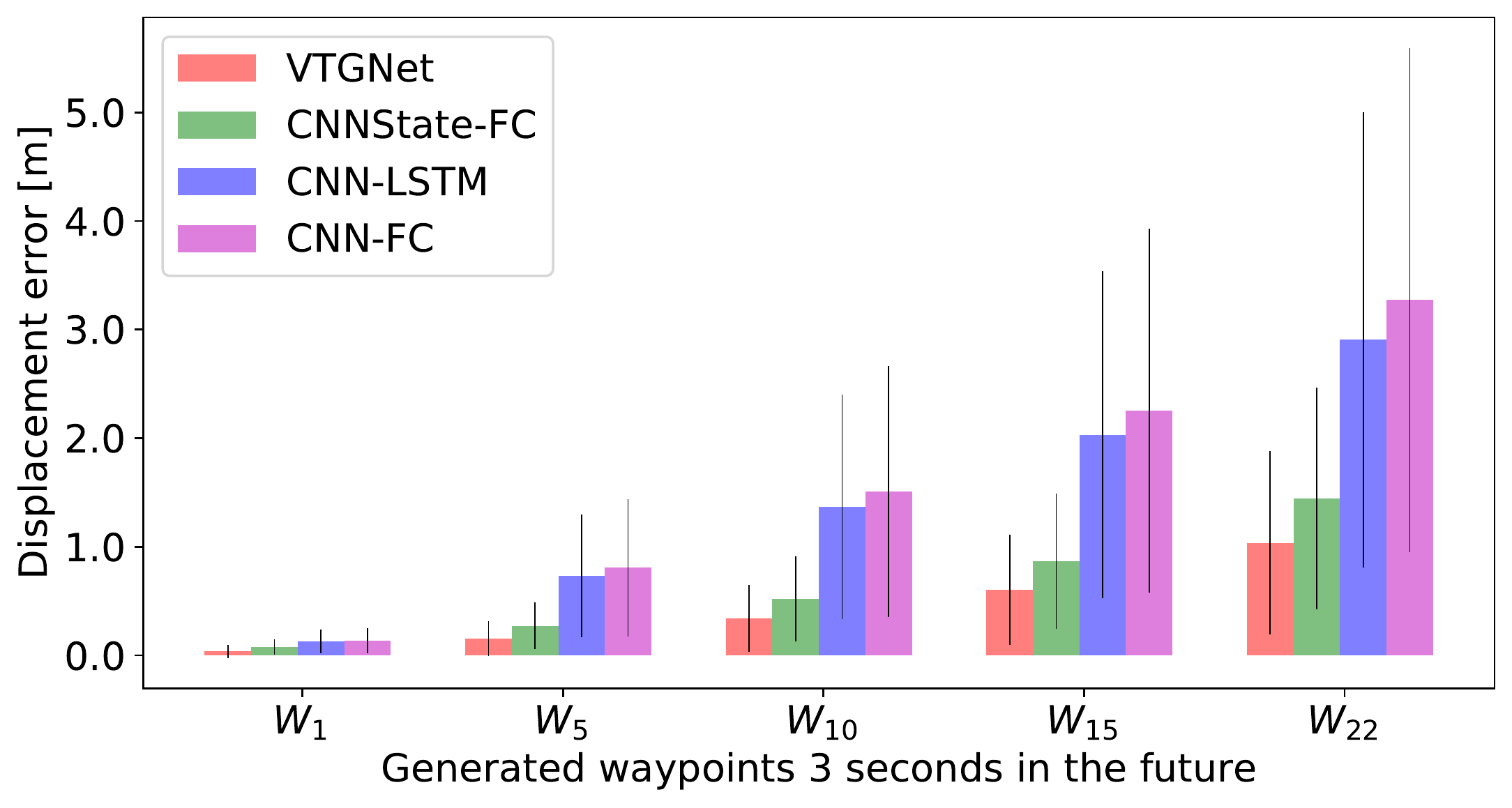}
        \setlength{\abovecaptionskip}{-0.3cm}
        \caption{Displacement error on the generated waypoints for different models. The black lines indicate the standard deviation.}
        \label{plot:baseline_results}     
\end{figure}

\subsection{Quantitative Analysis}

\subsubsection{Comparison to Baselines}
We first show the evaluation results on the test set from different model architectures in Table \ref{tab:quantitative}-\textsl{Models}. 

\textbf{Trajectory smoothness: }Table \ref{tab:quantitative} reveals that \texttt{CNN-FC} generates the smoothest trajectories with the lowest \textsl{Accel} value (0.181). \texttt{VTGNet} ranks second place (0.325). \texttt{CNN-LSTM} performs worst on this metric with the highest \textsl{Accel} (0.488).

\textbf{Error analyses: }From the second to the last column of Table \ref{tab:quantitative}, we can see that \texttt{VTGNet} generates more accurate trajectories than the other models in terms of the error-related metrics such as $\mathcal{E}_v$ and $\mathcal{E}_{ad}$. This superior performance indicates that our model performs more like humans than the other models. The displacement error over the 3.0 s preview horizon from different models is shown in Fig. \ref{plot:baseline_results}. Their errors and related variance all increase as the preview time increases, and our \texttt{VTGNet} achieves the best results with the smallest values. In addition, we can infer the reasonability of our network design by comparing the results from different baseline models, which are discussed as follows.

Table \ref{tab:quantitative} shows that both \texttt{CNN-LSTM} and \texttt{CNNState-FC} perform better than \texttt{CNN-FC} with less errors. The superiority of \texttt{CNN-LSTM} indicates that the recurrent architecture, such as LSTM, is much more efficient than the FC layers in processing spatiotemporal information. We conjecture that the LSTM module can infer the hidden state, such as environmental changes among the frames, which leads to better performance on this imitation learning task. However, it is difficult for the FC layers to extract valuable information directly from the consecutive visual features. On the other hand, the superiority of \texttt{CNNState-FC} indicates that past movement is beneficial for trajectory generation, which functions as the \textit{state memory}, similar to the memory of humans. Although the position and orientation information is contained in the raw images, using a CNN alone is not sufficient to extract this information and generate future trajectories.

\begin{table}
\newcommand{\tabincell}[2]{\begin{tabular}{@{}#1@{}}#2\end{tabular}}
        \renewcommand{\arraystretch}{1.3}
        \caption{The Driving Style Represented as the Mean and Standard Deviation of the Driving Acceleration $(m/s^2)$ for the Training Dataset and VTGNet.}
        \label{tab:driving_style}
        \centering
        \begin{tabular}{l  C{2.2cm} C{2.2cm}}
        \toprule[1pt]
        Environments & Training Data & VTGNet\\
        
        \midrule
        \texttt{\textit{Sun}} &0.38 $\pm$ 0.34 &0.40 $\pm$ 0.35 \\
        \texttt{\textit{Rain}} &0.39 $\pm$ 0.35 &0.29 $\pm$ 0.24 \\
        \texttt{\textit{Dusk}} &0.34 $\pm$ 0.31 &0.35 $\pm$ 0.32 \\
        \texttt{\textit{Night}} &0.32 $\pm$ 0.34 &0.32 $\pm$ 0.30 \\
        \texttt{\textit{Overcast}} &0.32 $\pm$ 0.33 &0.34 $\pm$ 0.29 \\
        \texttt{\textit{Snow}}  &0.21 $\pm$ 0.25 &0.26 $\pm$ 0.22 \\
        
        \bottomrule[1pt]

\end{tabular}
\end{table}

\begin{figure}[t]
        \centering 
        \includegraphics[width=\columnwidth]{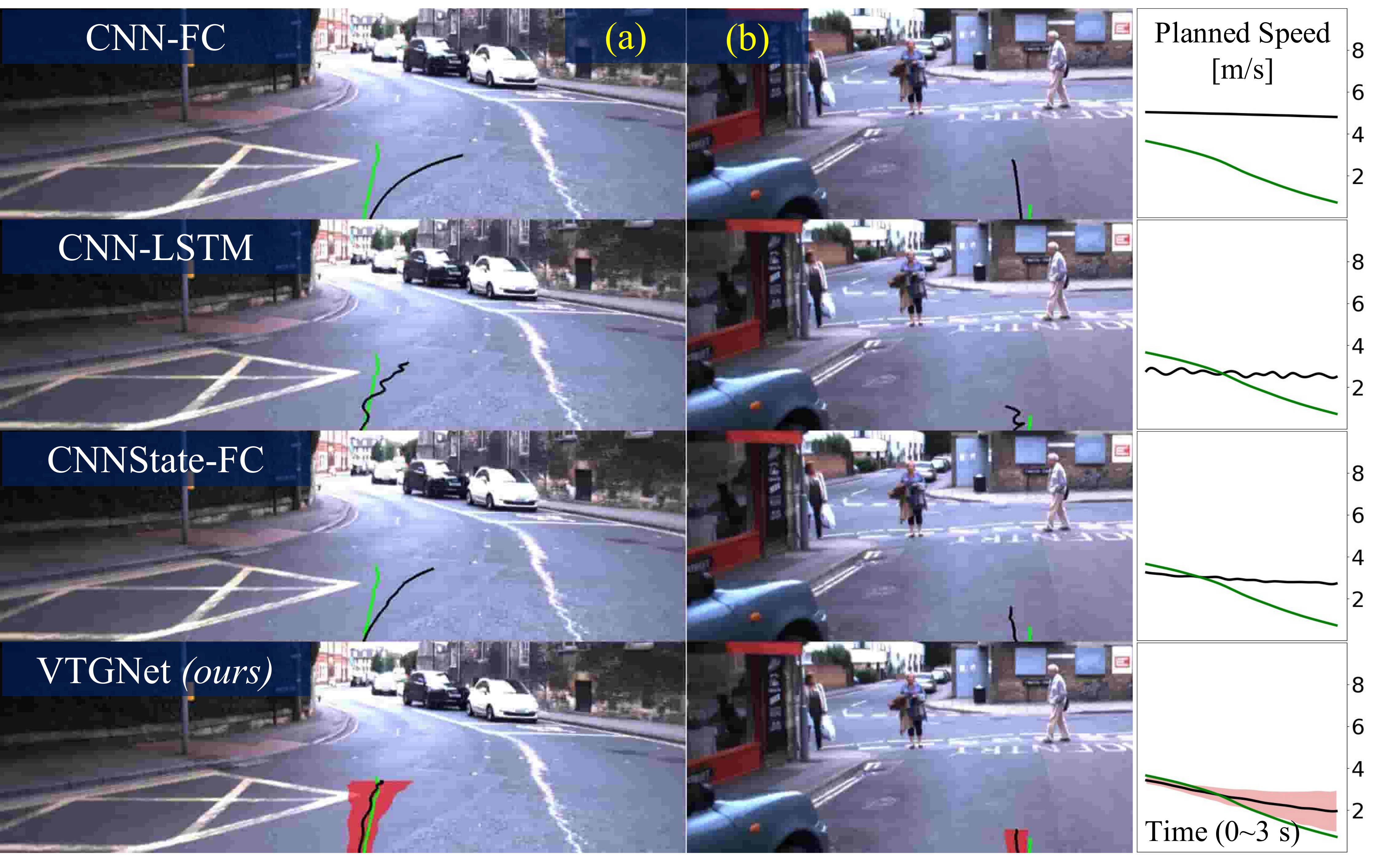}
        \caption{Sample results from different baselines and our \texttt{VTGNet}. The black lines indicate the planned results and the green lines indicate the ground truth. The red shaded areas represent the estimated uncertainty.}     
        \label{fig:baselines}     
\end{figure}

\subsubsection{Comparison among different environments}
\begin{figure*}[t]
        \centering 
        \includegraphics[width=2\columnwidth]{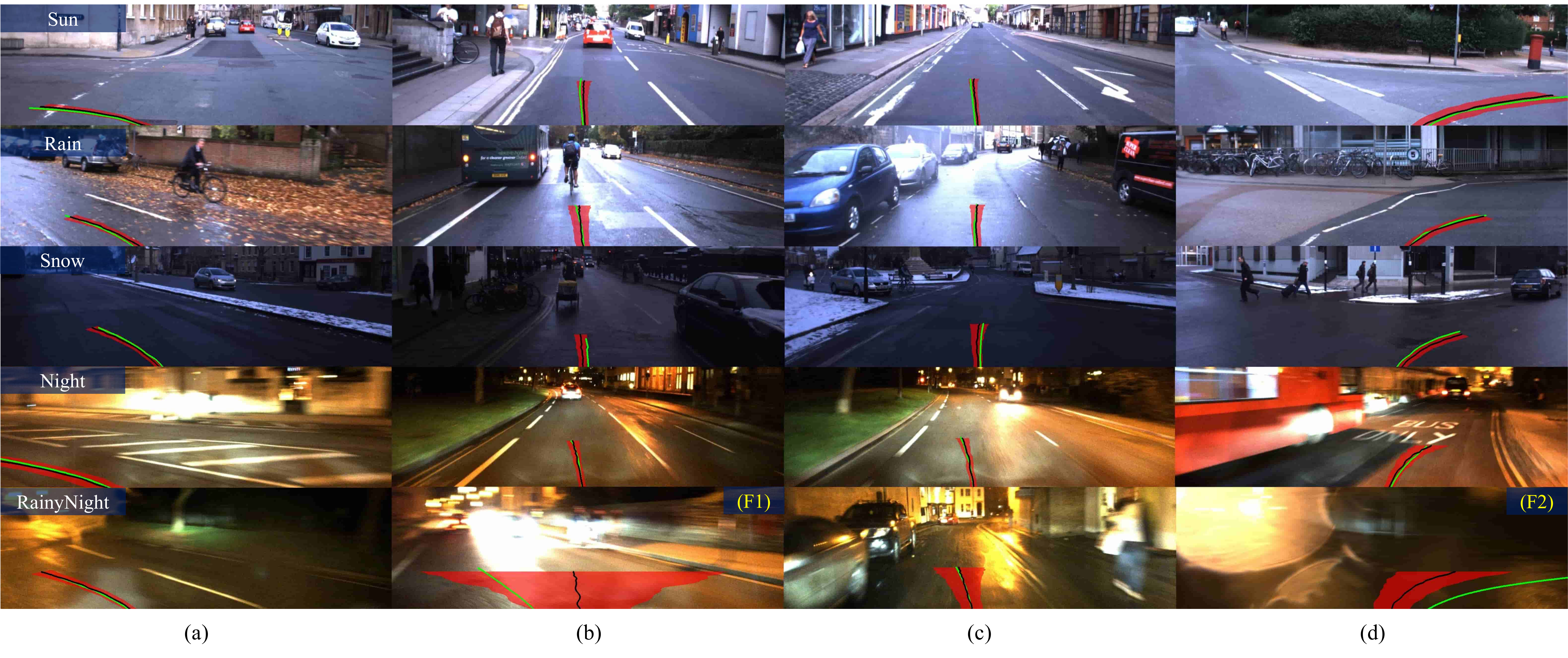}
        \setlength{\abovecaptionskip}{-0.15cm}
        \caption{Sample qualitative results for our model \texttt{VTGNet} in various environments. Column (a) shows the results for \textit{turn left}, and column (d) for \textit{turn right}. Column (b) shows the results for car following, and column (c) for lane keeping. The black lines indicate the planned results and the green lines indicate the ground truth. The red shaded areas represent the estimated uncertainty.}     
        \label{fig:qualitative}
\end{figure*}

In this section, we test and evaluate our \texttt{VTGNet} separately on different environments with various lighting and weather conditions, with the aim to see whether this architecture can generalize to different visual appearances. The results are shown in Table \ref{tab:quantitative}-\textsl{Environments}. It can be seen that \texttt{VTGNet} achieves robust results under different conditions and performs best in the \textit{snow} environment. In this scene, the generated trajectories have the lowest acceleration (0.262 $m/s^2$) and error results ($\mathcal{E}_v, \mathcal{E}_{ad}$, etc.). This is probably because the drivable areas are more distinct in this scenario with the undrivable areas covered by white snow. Another possible reason for the lower \textsl{Accel} value on snowy days is that the reference training data from humans reflects that they tend to drive with less dynamics on slippery roads in such weather, which affects our \texttt{VTGNet} trained with imitation learning. We verify this idea by quantitatively analyzing the driving styles of the training trajectories from humans and the generated trajectories from \texttt{VTGNet}. We formulate the driving style as the mean and standard deviation of the driving accelerations, and the results are shown in Table \ref{tab:driving_style}. It can be seen that in most conditions the driving styles of humans and \texttt{VTGNet} are quite similar, which accords with the principle of imitation learning.

\subsection{Qualitative Analysis}
\subsubsection{Comparison to Baselines} We compare \texttt{VTGNet} with the baselines, and the results are shown in Fig. \ref{fig:baselines}. It can be seen that although \texttt{CNN-FC} generates smooth trajectories, it deviates significantly from the ground truth values when taking turns in Fig. \ref{fig:baselines}-(a). In Fig. \ref{fig:baselines}-(b), the ego-vehicle should slow down for a pedestrian ahead, while \texttt{CNN-FC} maintains the current speed and moves forward. \texttt{CNN-LSTM} performs better in terms of errors, but the planned results from this model are rather jerky. In general, \texttt{VTGNet} achieves the best results.

\begin{figure}[t]
        \centering   
        \includegraphics[width=\columnwidth]{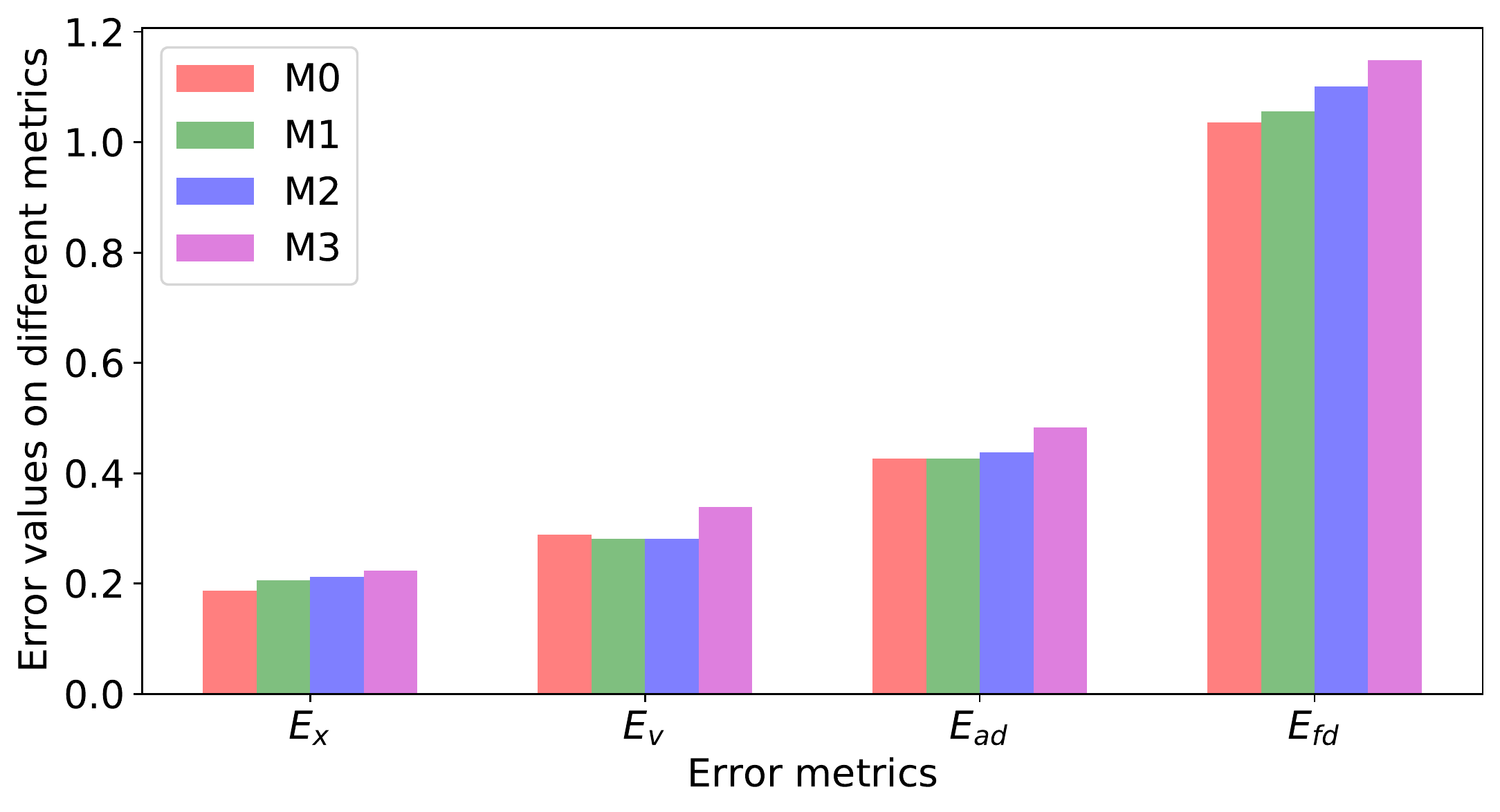}
        \caption{Error values of ablated models ($\mathcal{M}_1-\mathcal{M}_3$) and our \texttt{VTGNet} ($\mathcal{M}_0$). Lower values are better.}
        \label{plot:ablation}
\end{figure}

\begin{figure}[t]
        \centering 
        \includegraphics[width=\columnwidth]{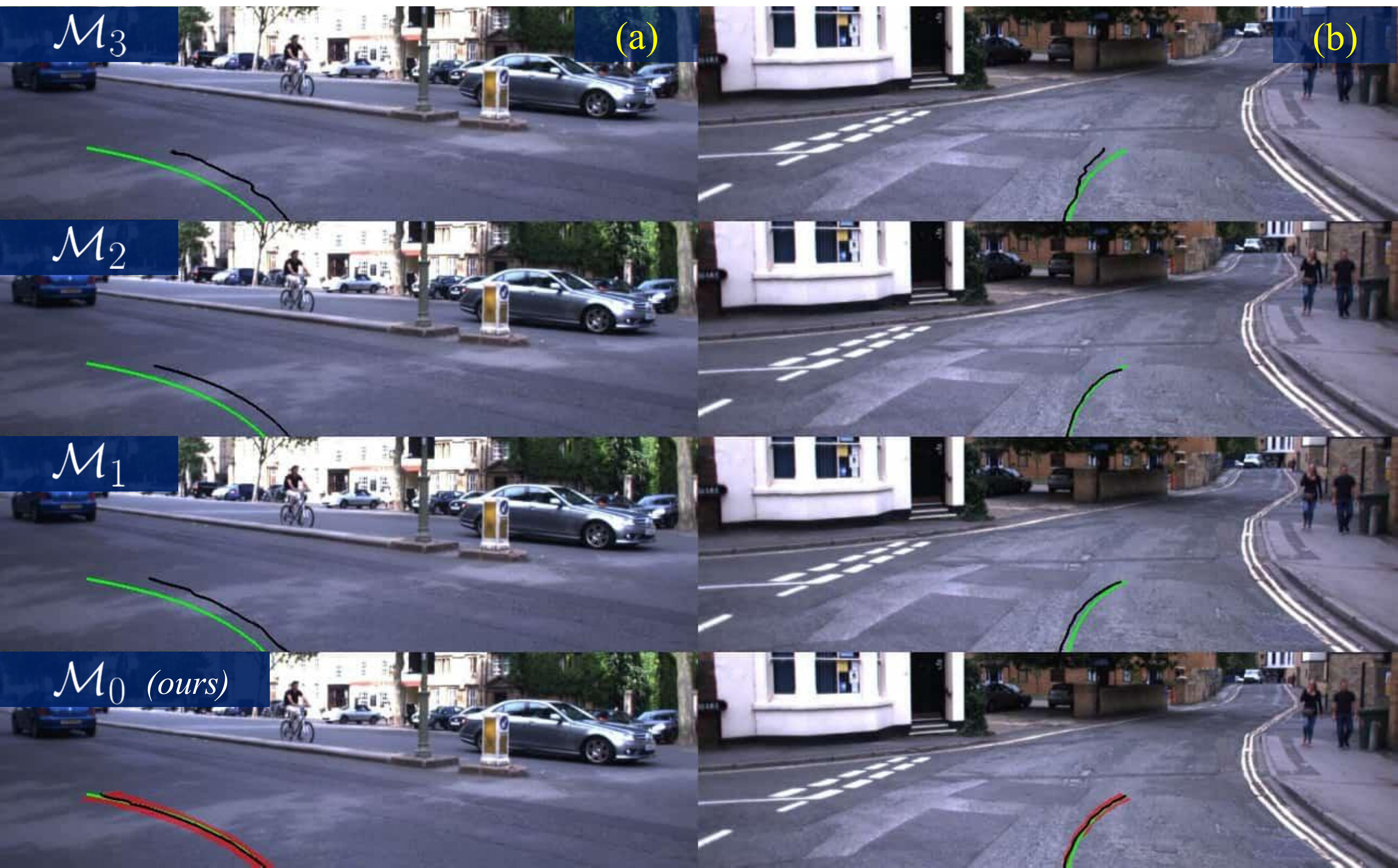}
        \caption{Sample results from ablated models ($\mathcal{M}_1-\mathcal{M}_3$) and our \texttt{VTGNet} ($\mathcal{M}_0$). The black lines indicate the planned results and the green lines indicate the ground truth. The red shaded areas represent the estimated uncertainty.}     
        \label{fig:ablation}     
\end{figure}

\subsubsection{Results in different environments}
Figure \ref{fig:qualitative} displays sample qualitative results of \texttt{VTGNet} in 5 typical environments, in which the challenging scene \textit{rainynight} is presented in the last row. Because of the raindrops on the camera lens, the halo effects in this scene are sometimes conspicuous. It is worth noting that this scene is not included in the training set, and is only used to qualitatively test the generalization performance. In general, we can see that our \texttt{VTGNet} is able to generate collision-free trajectories under various lighting and weather conditions. The performance is also reliable in some tough scenarios. For example, in \textit{night}-(d), the bus in the left area presents a severe motion artifact, and \texttt{VTGNet} generates a safe trajectory to turn right. In \textit{rainynight}-(c), the motion artifact is more severe but the generated results are still reasonable and similar to the ground truth (with higher uncertainty). We consider that the well-balanced training dataset of high diversity in terms of lighting and weather conditions contributes to such generalizable performance.

Additionally, we show some failure cases, indicated by extreme uncertainties in \textit{rainynight}-(b,d), where the input image is heavily noised. In such scenarios, the system should drive the vehicle more cautiously to avoid potential accidents.

\subsection{Ablation Studies}
We train and test a series of ablated models $\mathcal{M}_1 \sim \mathcal{M}_3$ to show the benefits of the \texttt{VTGNet} design ($\mathcal{M}_0$). $\mathcal{M}_1$ removes the uncertainty estimation from $\mathcal{M}_0$. $\mathcal{M}_2$ further removes the self-attention module from $\mathcal{M}_1$. $\mathcal{M}_3$ is a variant of $\mathcal{M}_2$, which adopts two LSTM modules to process image and movement sequences, separately. We show the error values from these models in Fig. \ref{plot:ablation}. In general, the performance of the model decreases from $\mathcal{M}_0$ to $\mathcal{M}_3$, indicating the rationality of our \texttt{VTGNet}. The superiority of $\mathcal{M}_0$ over $\mathcal{M}_1$ is consistent with the finding in \cite{Kendall2017WhatUD} where modeling data uncertainty improves prediction performance. On the other hand, $\mathcal{M}_3$ performs worse than $\mathcal{M}_2$, from which we can conclude that using a single LSTM module to process the \textit{multi-modal} information together would help to generate more reasonable outputs. Furthermore, we show a qualitative comparison in Fig. \ref{fig:ablation}, where the results from $\mathcal{M}_0$ are both accurate and smooth.

\begin{figure}[t]
        \centering   
        \includegraphics[width=\columnwidth]{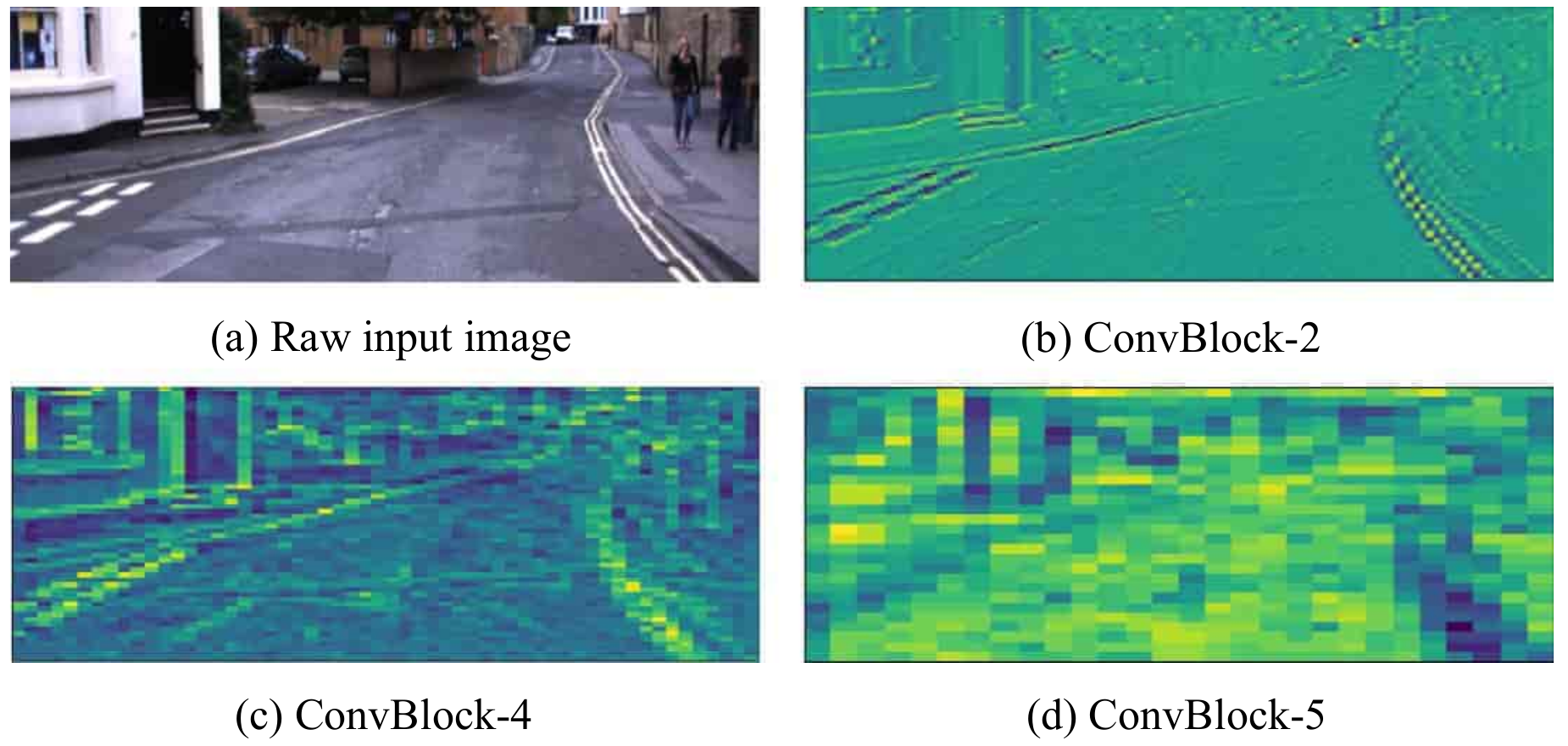}
        \caption{Image feature visualization of our \texttt{VTGNet}. (a) The input RGB image. (b-d) The averaged feature maps of some bottleneck convolutional layers in the feature extractor.}     
        \label{fig:feature_map}
\end{figure}

\subsection{Discussion and Reasoning}
\subsubsection{Feature Extractor Analysis}
Compared with hand-crafted feature extractors, one advantage of CNNs is that they can learn to extract useful features automatically from raw RGB images. We confirm this idea by visualizing the averaged feature maps from a series of convolution layers of our model. The results are shown in Fig. \ref{fig:feature_map}. We can see that the obvious extracted features are the road boundaries, which are helpful for generating the trajectories.

\subsubsection{Attention Analysis}
\begin{figure}[t]
        \centering   
        \includegraphics[width=\columnwidth]{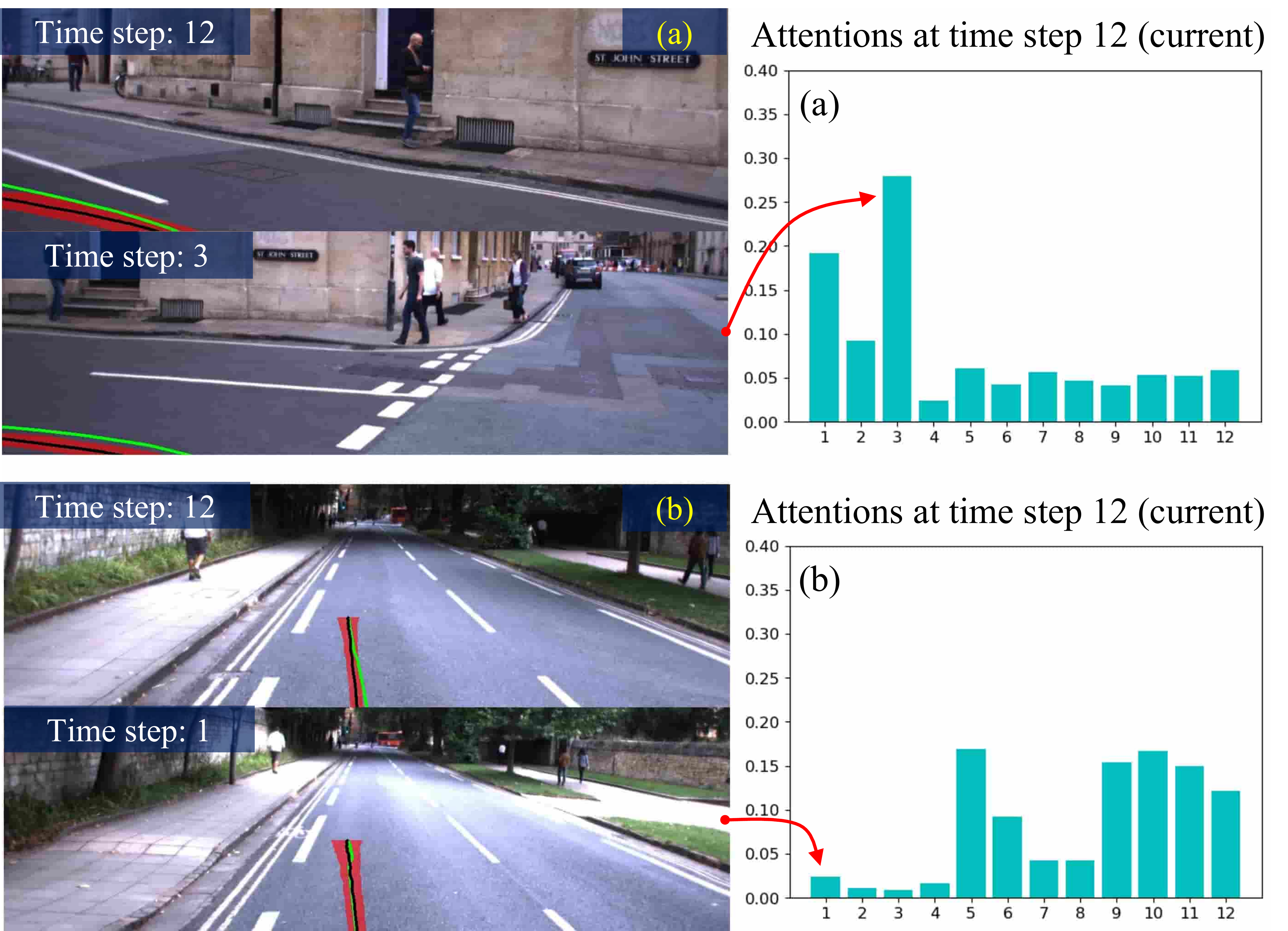}
        \caption{Attention distribution from our \texttt{VTGNet} in two typical scenarios. (a) Turning at intersections. (b) Lane following.}
        \label{fig:attention}
\end{figure}

Here, we analyze the attention mechanism of our model using Fig. \ref{fig:attention}. We can see that when taking turns (Fig. \ref{fig:attention}-(a)), the observations at intersection entrances (time step 3) are weighted with more attentions, which contains more structural information of the surroundings. On the other hand, when following lanes on roads, more recent observations are assigned with more attention because the past observations are rather redundant, providing no extra valuable information.

\section{Closed-Loop Experiments in the CARLA Driving Simulator}
In this section, we aim to determine whether our \texttt{VTGNet} can perform more complicated tasks, such as recovering from mistakes and driving in dynamic traffic. To this end, we conduct closed-loop experiments in the CARLA driving simulator\cite{dosovitskiy2017carla} to further show its performance. CARLA is an open-source simulator providing a high-fidelity dynamic world and different vehicles with realistic physics.

\begin{table*}
\newcommand{\tabincell}[2]{\begin{tabular}{@{}#1@{}}#2\end{tabular}}
        \renewcommand{\arraystretch}{1.3}
        \caption{Closed-loop Evaluation Results on Success Rate (\%) in the Carla Driving Simulator.}
        \label{tab:carla_evaluation}
        \centering
        \begin{tabular}{l C{1.3cm} C{1.3cm}  C{1.3cm} C{1.3cm}  C{1.3cm} C{1.3cm}  }
        \toprule[1pt]
        {}&
        \multicolumn{2}{c}{{Training Conditions}}&
        \multicolumn{2}{c}{{New Vehicle}}&
        \multicolumn{2}{c}{{New Vehicle \& Town}}\\
        \cmidrule(lr){2-3} \cmidrule(lr){4-5} \cmidrule(lr){6-7}
        Traffic & Empty & Dynamic & Empty & Dynamic & Empty & Dynamic\\
        \midrule
        \texttt{CILRS}\cite{codevilla2019exploring} & 1.7 & 3.3 & 8.3  & 3.3 & 10.0 & 3.3  \\
        \texttt{CILRS++} & \textbf{95.0} & \textbf{86.7} & 15.0  & 13.3 & 23.3 & 25.0  \\
        \texttt{VTGNet} & 93.3 &75.0 & \textbf{43.3}  &\textbf{40.0} & \textbf{35.0} & \textbf{33.3}  \\
        
        \bottomrule[1pt]

\end{tabular}
\end{table*}

\begin{figure*}[t]
        \centering   
        \includegraphics[width=2\columnwidth]{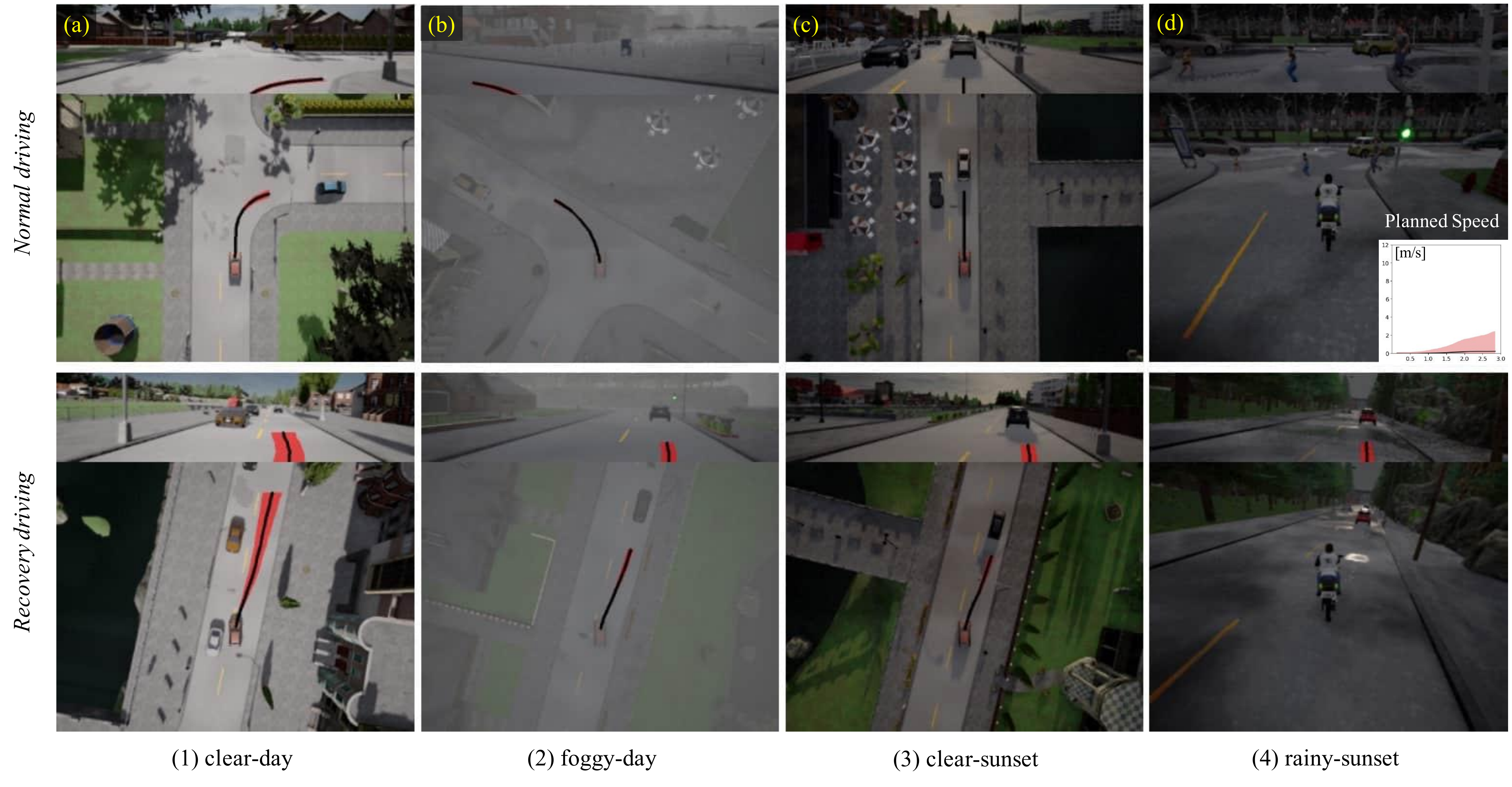}
        \caption{Closed-loop evaluation results of \texttt{VTGNet} in our \textit{AddNoise} benchmark. Row-1 shows some normal driving behaviors: (a,b) turning at intersections, (c) vehicle-following and (d) stopping for pedestrians. Row-2 shows recovery behaviors against off-center/orientation errors under different weather conditions. The black lines indicate the planned results and the red shaded areas represent the estimated uncertainty.}
        \label{fig:carla}
\end{figure*}

\subsection{Implementation Details}
\subsubsection{Domain Adaption to New Environments}
\label{subsubsec:domain_adaotion}
Our prior model is trained with the Robotcar dataset, which lacks reference trajectories to recover from off-center or off-orientation mistakes. On the other hand, the domain gap in terms of visual difference also prevents us from directly deploying the model in CARLA. To improve the performance, we first collect a new expert driving dataset in CARLA, named VTG-Driving, and then refine our model with this dataset to achieve \textit{domain adaption}. During data collection, we set random routes ranging from 300 m to 1500 m in Town01 of CARLA and drive the vehicle at the desired speed of 40 km/h. The direction command is given by a high-level global planner of CARLA. To construct a dynamic environment, we set five weather conditions: \textit{clear-day}, \textit{clear-sunset}, \textit{foggy-day}, \textit{rainy-day}, and \textit{rainy-sunset}. For each weather condition, we run 100 episodes with dynamic roaming pedestrians and vehicles that are controlled by the AI engine of CARLA. Furthermore, we add random \textit{steering} noise to the vehicle every 6 s to collect error-recovery trajectories. The final dataset lasts 16.6 hours and covers a driving distance of 288.7 km, which is also released for public research.

\subsubsection{Evaluation Benchmark}
The prior benchmarks of CARLA mainly concentrate on the ability to handle complex visual appearances \cite{dosovitskiy2017carla} or the ability to drive in urban traffic with dynamic obstacles \cite{codevilla2019exploring}. We extend these ideas and propose a much tougher benchmark \textit{AddNoise} to also evaluate the ability to recover from off-center/orientation errors. Similar to the data collection process, we add random \textit{steering} noise to the test vehicle every 5 s that lasts 0.2$\sim$1.0 s. Moreover, we examine the cross-platform ability by changing the training \textit{car} to an unseen \textit{motorcycle}, which is smaller in size and has a smaller turning radius and larger acceleration.

In this benchmark, the test vehicle should drive in empty and dynamic (with other dynamic vehicles/pedestrians) environments against the noise introduced above. Each task corresponds to 60 goal-directed episodes under the 5 weather conditions introduced in Section \ref{subsubsec:domain_adaotion}. In each episode, the vehicle starts from a random position and is directed by a high-level planner to reach the destination. The episode is considered to be successful if the vehicle arrives at the destination within the time limit and does not collide with other objects. Finally, the success rate is used to measure the autonomous driving ability.

\subsubsection{Baselines for Comparison}
We have shown the superiority of \texttt{VTGNet} in the area of end-to-end trajectory \textit{planning} in Section \ref{sec:experiments_robotcar}. In this section, we further compare it with a recent SOTA vision-based end-to-end \textit{control} network named \texttt{CILRS}\cite{codevilla2019exploring}. This model is trained with data unaffected by noise (filtered from VTG-Driving) following the setup as detailed in the original paper. We also train another model named \texttt{CILRS++} on our full dataset including the noise-recovery behaviors. The reference expert actions for training are the original actions that are not overlaid with noise. Note that we replace the original image processing backbone of these two models to MobileNet-V2 as our \texttt{VTGNet} for fair comparison. For \texttt{VTGNet}, we design two PID controllers for the test vehicles (car/motorcycle) to translate the generated trajectory into driving actions.

\subsection{Results}
\subsubsection{Quantitative Analysis}
The evaluation results are shown in Table \ref{tab:carla_evaluation}. Although \texttt{CILRS} achieves good performance in \cite{codevilla2019exploring}, with a reported success rate of 42\%$\sim$97\% in training conditions, its performance degrades significantly in \textit{AddNoise} and the success rate is only 1.7\%$\sim$3.3\%. After being augmented with training data that recovers from mistakes, \texttt{CILRS++} performs much better against the noisy test environment and achieves higher success rate of 86.7\%$\sim$95.0\% in training conditions. However, these two models performs badly on new platforms, i.e., a motorcycle in this work, and the highest success rate is only 25.0\%. The reason is that the learned driving policies of end-to-end control systems can only perform well on data collected with specifically calibrated actuation setups, which also accords with the statements in \cite{bddv}. Actually, we observe a very shaky and unstable driving performance of these two models where the motorcycle often rushes to the roadside and collides with static objects such as lamps and fences. Note that the new test town (Town02) in CARLA contains less static objects on the roadside than Town01, and therefore the results for \textit{New Vehicle\&Town} are better than for \textit{New Vehicle} for \texttt{CILRS++}.

On the other hand, our \texttt{VTGNet} performs a little worse than \texttt{CILRS++} in training conditions (75.0\%$\sim$93.3\%) but achieves much better generalization results on new vehicle and new town. For example, the success rate of our model in dynamic environments of the $New Vehicle$ setup is 40.0\%, which is three times higher than that of \texttt{CILRS++}. We accredit such superiority to our model architecture, which allows the control module to flexibly adapt to different vehicle platforms. Since the motorcycle has smaller turning angles, it is more susceptible to the \textit{steering} noises than the \textit{car} is for \textit{Training Conditions} in our benchmark. Because of this, very large off-center and off-orientation errors may occur after the noise process. For example, the motorcycle is often disturbed onto sidewalks where the road is out of sight, which is hard to recover and explains the performance degradation from \textit{Training Conditions} to \textit{New Vehicle}.

\subsubsection{Qualitative Analysis}
We show some driving snapshots of \texttt{VTGNet} in Fig. \ref{fig:carla}. It shows that \texttt{VTGNet} not only adapts to different weather conditions but also possesses the ability to perform reactive driving such as following a vehicle (Fig. \ref{fig:carla}-(c)) and stopping for pedestrians (Fig. \ref{fig:carla}-(d)). Furthermore, we show its recovery behaviors in Fig. \ref{fig:carla}-(b), where the ego-car is autonomously recovering from off-center/orientation errors disturbed by periodic steering noises. More related results are shown in the supplemental videos\footnote{\url{https://sites.google.com/view/vtgnet/}}.

\begin{figure}[t]
        \centering   
        \includegraphics[width=\columnwidth]{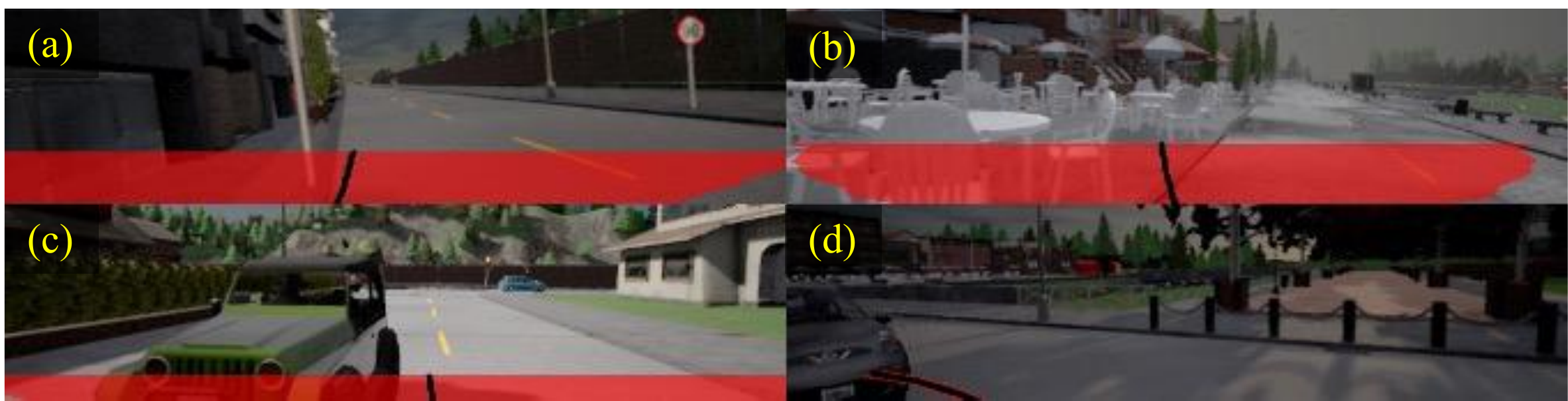}
        \caption{Failure cases of our driving model. (a-c) Most failure cases (80\%) are informed with a high uncertainty output, which can be avoided with timely human intervention. (d) A few failure cases can not be captured with the estimated uncertainty. The black lines indicate the planned results and the red shaded areas represent the estimated uncertainty.}
        \label{fig:carla_failure}
\end{figure}

\subsubsection{Benefits and Limitations of Uncertainty Estimation for Safe Driving}
We observe that 80\% of failure cases in the benchmark results of \texttt{VTGNet} are informed with a high uncertainty output, and thus can be avoided by timely human intervention, as shown in Fig. \ref{fig:carla_failure}-(a,b,c). However, robust functional safety measures are still beyond reach because in a few cases, the model is too confident in its wrong prediction with low uncertainty, as shown in Fig. \ref{fig:carla_failure}-(d), where the ego-vehicle collides with the vehicle ahead without slowing down. Such limitations leave us possible avenues for future research on safer autonomous driving.

\section{Conclusion and Future Work}

In this paper, to improve the driving ability of current learning-based methods in terms of safety measures and recovery behaviors against errors, we proposed an uncertainty-aware vision-based trajectory generation network for AVs, named VTGNet. For acceptable training and testing performance, we first created a large-scale driving dataset from the original Robotcar dataset. Then, we implemented the proposed network with a twofold pipeline. The first part is a feature extractor composed of bottleneck CNN layers based on MobileNet-V2. The second part is a trajectory generator, which consists of a self-attention module to weight the redundant history information, and an LSTM module to process the spatiotemporal features. The overall framework was designed to take as input the front-view image and movement sequences in the past 1.5 seconds with the intention command to generate feasible trajectories 3 seconds in the future. The whole network is differentiable and was then trained end-to-end using imitation learning. Third, we validated the robustness and reliability of our proposed network under various weather and lighting conditions by comparing it with 3 baselines and several ablated models. Finally, we conducted extensive closed-loop experiments in the CARLA driving simulator to demonstrate the more advanced ability of VTGNet to recover from off-center/orientation errors and cross-platform driving. Furthermore, we showed that most of the failure cases can be avoided by timely human-intervention with the estimated uncertainty, which is beneficial to develop safety-critical driving systems.

Despite the success of the proposed network, there still exist some limitations of our model. 1) Robust safety measures can not be guaranteed even with the uncertainty estimation. More studies are needed in this area to develop safer autonomous driving systems. 2) The discrete high-level driving commands are not suitable for more complex road topologies such as intersections with more than one possibility to turn. It is expected to incorporate global route information to provide more detailed driving intentions into the network. 3) Camera images may not be reliable under extreme lighting conditions, as shown in Fig. \ref{fig:qualitative}-(F1,F2). We cannot expect the system to generate acceptable trajectories with such input data. Therefore, in the future, we aim to develop a framework based on our current work to use multi-modal information from complementary sensors such as lidar, radar and a thermal camera. We believe that by doing so, the system could be more robust in challenging environments.

\bibliographystyle{IEEEtran}
\bibliography{ref.bib}

\begin{thebibliography}{10}
\providecommand{\url}[1]{#1}
\csname url@samestyle\endcsname
\providecommand{\newblock}{\relax}
\providecommand{\bibinfo}[2]{#2}
\providecommand{\BIBentrySTDinterwordspacing}{\spaceskip=0pt\relax}
\providecommand{\BIBentryALTinterwordstretchfactor}{4}
\providecommand{\BIBentryALTinterwordspacing}{\spaceskip=\fontdimen2\font plus
\BIBentryALTinterwordstretchfactor\fontdimen3\font minus
  \fontdimen4\font\relax}
\providecommand{\BIBforeignlanguage}[2]{{%
\expandafter\ifx\csname l@#1\endcsname\relax
\typeout{** WARNING: IEEEtran.bst: No hyphenation pattern has been}%
\typeout{** loaded for the language `#1'. Using the pattern for}%
\typeout{** the default language instead.}%
\else
\language=\csname l@#1\endcsname
\fi
#2}}
\providecommand{\BIBdecl}{\relax}
\BIBdecl

\bibitem{global_report1}
W.~H. Organization \emph{et~al.}, \emph{Global status report on alcohol and
  health-2014}.\hskip 1em plus 0.5em minus 0.4em\relax World Health
  Organization, 2014.

\bibitem{human_error}
S.~Singh, ``Critical reasons for crashes investigated in the national motor
  vehicle crash causation survey,'' 2015.

\bibitem{global_report3}
A.~Brown, B.~Repac, and J.~Gonder, ``Autonomous vehicles have a wide range of
  possible energy impacts,'' NREL, University of Maryland, Tech. Rep., 2013.

\bibitem{fan2020sne}
R.~Fan, H.~Wang, P.~Cai, and M.~Liu, ``Sne-roadseg: Incorporating surface
  normal information into semantic segmentation for accurate freespace
  detection,'' in \emph{European Conference on Computer Vision}.\hskip 1em plus
  0.5em minus 0.4em\relax Springer, 2020, pp. 340--356.

\bibitem{kuutti2020survey}
S.~{Kuutti}, R.~{Bowden}, Y.~{Jin}, P.~{Barber}, and S.~{Fallah}, ``A survey of
  deep learning applications to autonomous vehicle control,'' \emph{IEEE
  Transactions on Intelligent Transportation Systems}, pp. 1--22, 2020.

\bibitem{survey_trajectory}
B.~Paden, M.~{\v{C}}{\'a}p, S.~Z. Yong, D.~Yershov, and E.~Frazzoli, ``A survey
  of motion planning and control techniques for self-driving urban vehicles,''
  \emph{IEEE Transactions on intelligent vehicles}, vol.~1, no.~1, pp. 33--55,
  2016.

\bibitem{schwarting2018planning}
W.~Schwarting, J.~Alonso-Mora, and D.~Rus, ``Planning and decision-making for
  autonomous vehicles,'' \emph{Annual Review of Control, Robotics, and
  Autonomous Systems}, 2018.

\bibitem{cai2019vision}
P.~{Cai}, Y.~{Sun}, Y.~{Chen}, and M.~{Liu}, ``Vision-based trajectory planning
  via imitation learning for autonomous vehicles,'' in \emph{2019 IEEE
  Intelligent Transportation Systems Conference (ITSC)}, Oct 2019, pp.
  2736--2742.

\bibitem{mcallister2017concrete}
R.~McAllister, Y.~Gal, A.~Kendall, M.~van~der Wilk, A.~Shah, R.~Cipolla, and
  A.~Weller, ``Concrete problems for autonomous vehicle safety: Advantages of
  {Bayesian} deep learning,'' in \emph{IJCAI 2017}, 2017.

\bibitem{learn_to_drive}
A.~Kendall, J.~Hawke, D.~Janz, P.~Mazur, D.~Reda, J.-M. Allen, V.-D. Lam,
  A.~Bewley, and A.~Shah, ``Learning to drive in a day,'' \emph{arXiv preprint
  arXiv:1807.00412}, 2018.

\bibitem{yurtsever2019survey}
E.~Yurtsever, J.~Lambert, A.~Carballo, and K.~Takeda, ``A survey of autonomous
  driving: common practices and emerging technologies,'' \emph{arXiv preprint
  arXiv:1906.05113}, 2019.

\bibitem{eed1}
M.~Bojarski, D.~Del~Testa, D.~Dworakowski, B.~Firner, B.~Flepp, P.~Goyal, L.~D.
  Jackel, M.~Monfort, U.~Muller, J.~Zhang \emph{et~al.}, ``End to end learning
  for self-driving cars,'' \emph{arXiv preprint arXiv:1604.07316}, 2016.

\bibitem{eed2}
J.~Jhung, I.~Bae, J.~Moon, T.~Kim, J.~Kim, and S.~Kim, ``End-to-end steering
  controller with {CNN-based} closed-loop feedback for autonomous vehicles,''
  in \emph{2018 IEEE Intelligent Vehicles Symposium (IV)}.\hskip 1em plus 0.5em
  minus 0.4em\relax IEEE, 2018, pp. 617--622.

\bibitem{cai2020probabilistic}
P.~Cai, S.~Wang, Y.~Sun, and M.~Liu, ``Probabilistic end-to-end vehicle
  navigation in complex dynamic environments with multimodal sensor fusion,''
  \emph{IEEE Robotics and Automation Letters}, vol.~5, no.~3, pp. 4218--4224,
  2020.

\bibitem{condition-eed}
F.~Codevilla, M.~Miiller, A.~L{\'o}pez, V.~Koltun, and A.~Dosovitskiy,
  ``End-to-end driving via conditional imitation learning,'' in \emph{2018 IEEE
  International Conference on Robotics and Automation (ICRA)}.\hskip 1em plus
  0.5em minus 0.4em\relax IEEE, 2018, pp. 1--9.

\bibitem{bddv}
H.~Xu, Y.~Gao, F.~Yu, and T.~Darrell, ``End-to-end learning of driving models
  from large-scale video datasets,'' in \emph{Proceedings of the IEEE
  conference on computer vision and pattern recognition}, 2017, pp. 2174--2182.

\bibitem{Mller2018DrivingPT}
M.~M{\"u}ller, A.~Dosovitskiy, B.~Ghanem, and V.~Koltun, ``Driving policy
  transfer via modularity and abstraction,'' in \emph{CoRL}, 2018.

\bibitem{codevilla2019exploring}
F.~Codevilla, E.~Santana, A.~M. L{\'o}pez, and A.~Gaidon, ``Exploring the
  limitations of behavior cloning for autonomous driving,'' in
  \emph{Proceedings of the IEEE International Conference on Computer Vision},
  2019, pp. 9329--9338.

\bibitem{thesis}
M.~Bergqvist and O.~R\"{o}dholm, ``{Deep Path Planning Using Images and Object
  Data},'' Master's thesis, Chalmers University of Technology, Gothenburg,
  Sweden, 2018.

\bibitem{r2p2}
N.~Rhinehart, K.~M. Kitani, and P.~Vernaza, ``{R2P2}: A reparameterized
  pushforward policy for diverse, precise generative path forecasting,'' in
  \emph{Proceedings of the European Conference on Computer Vision (ECCV)},
  2018, pp. 772--788.

\bibitem{waymo}
M.~Bansal, A.~Krizhevsky, and A.~Ogale, ``Chauffeurnet: Learning to drive by
  imitating the best and synthesizing the worst,'' \emph{arXiv preprint
  arXiv:1812.03079}, 2018.

\bibitem{alvinn}
D.~A. Pomerleau, ``Alvinn: An autonomous land vehicle in a neural network,'' in
  \emph{Advances in neural information processing systems}, 1989, pp. 305--313.

\bibitem{wu2019end}
T.~Wu, A.~Luo, R.~Huang, H.~Cheng, and Y.~Zhao, ``End-to-end driving model for
  steering control of autonomous vehicles with future spatiotemporal
  features,'' in \emph{2019 IEEE/RSJ International Conference on Intelligent
  Robots and Systems (IROS)}.\hskip 1em plus 0.5em minus 0.4em\relax IEEE,
  2019, pp. 950--955.

\bibitem{cai2020high}
P.~Cai, X.~Mei, L.~Tai, Y.~Sun, and M.~Liu, ``High-speed autonomous drifting
  with deep reinforcement learning,'' \emph{IEEE Robotics and Automation
  Letters}, vol.~5, no.~2, pp. 1247--1254, 2020.

\bibitem{eth-route}
S.~Hecker, D.~Dai, and L.~Van~Gool, ``End-to-end learning of driving models
  with surround-view cameras and route planners,'' in \emph{Proceedings of the
  European Conference on Computer Vision (ECCV)}, 2018, pp. 435--453.

\bibitem{yang2018scene}
S.~Yang, W.~Wang, C.~Liu, and W.~Deng, ``Scene understanding in deep
  learning-based end-to-end controllers for autonomous vehicles,'' \emph{IEEE
  Transactions on Systems, Man, and Cybernetics: Systems}, vol.~49, no.~1, pp.
  53--63, 2018.

\bibitem{bahdanau2014neural}
D.~Bahdanau, K.~Cho, and Y.~Bengio, ``Neural machine translation by jointly
  learning to align and translate,'' \emph{arXiv preprint arXiv:1409.0473},
  2014.

\bibitem{zhang2019heterogeneous}
C.~Zhang, D.~Song, C.~Huang, A.~Swami, and N.~V. Chawla, ``Heterogeneous graph
  neural network,'' in \emph{Proceedings of the 25th ACM SIGKDD International
  Conference on Knowledge Discovery \& Data Mining}, 2019, pp. 793--803.

\bibitem{Kendall2017WhatUD}
A.~Kendall and Y.~Gal, ``What uncertainties do we need in bayesian deep
  learning for computer vision?'' in \emph{NIPS}, 2017.

\bibitem{Tai2019VisualbasedAD}
L.~Tai, P.~Yun, Y.~Chen, C.~Liu, H.~Ye, and M.~Liu, ``Visual-based autonomous
  driving deployment from a stochastic and uncertainty-aware perspective,''
  \emph{2019 IEEE/RSJ International Conference on Intelligent Robots and
  Systems (IROS)}, pp. 2622--2628, 2019.

\bibitem{Fan2019LearningRB}
T.~Fan, P.~Long, W.~Liu, J.~Pan, R.~Yang, and D.~Manocha, ``Learning resilient
  behaviors for navigation under uncertainty environments,'' \emph{ArXiv}, vol.
  abs/1910.09998, 2019.

\bibitem{time_ref}
M.~Green, ``" how long does it take to stop?" methodological analysis of driver
  perception-brake times,'' \emph{Transportation human factors}, vol.~2, no.~3,
  pp. 195--216, 2000.

\bibitem{mobilenetv2}
M.~Sandler, A.~Howard, M.~Zhu, A.~Zhmoginov, and L.-C. Chen, ``Mobilenetv2:
  Inverted residuals and linear bottlenecks,'' in \emph{Proceedings of the IEEE
  Conference on Computer Vision and Pattern Recognition}, 2018, pp. 4510--4520.

\bibitem{intention}
W.~Gao, D.~Hsu, W.~S. Lee, S.~Shen, and K.~Subramanian, ``Intention-net:
  Integrating planning and deep learning for goal-directed autonomous
  navigation,'' \emph{arXiv preprint arXiv:1710.05627}, 2017.

\bibitem{adam}
D.~P. Kingma and J.~Ba, ``Adam: A method for stochastic optimization,''
  \emph{arXiv preprint arXiv:1412.6980}, 2014.

\bibitem{metrics}
S.~Pellegrini, A.~Ess, K.~Schindler, and L.~Van~Gool, ``You'll never walk
  alone: Modeling social behavior for multi-target tracking,'' in
  \emph{Computer Vision, 2009 IEEE 12th International Conference on}.\hskip 1em
  plus 0.5em minus 0.4em\relax IEEE, 2009, pp. 261--268.

\bibitem{dosovitskiy2017carla}
A.~Dosovitskiy, G.~Ros, F.~Codevilla, A.~Lopez, and V.~Koltun, ``{CARLA}: {An}
  open urban driving simulator,'' in \emph{Proceedings of the 1st Annual
  Conference on Robot Learning}, 2017, pp. 1--16.

\end{thebibliography}

\end{document}